\documentclass[preprint,12pt]{elsarticle}

\usepackage{graphicx}
\usepackage{amssymb}

\usepackage{lineno}

\usepackage{graphicx}
\usepackage[utf8]{inputenc}
\usepackage{lmodern}
\usepackage[T1]{fontenc}
\usepackage{lscape}
\usepackage{amsmath}	
\usepackage{amssymb}
\usepackage{cases}
\usepackage{graphicx}
\usepackage{caption}
\usepackage{comment}

\usepackage{makecell}
\usepackage{multirow}
\usepackage{amsfonts}
\usepackage{float} 
\usepackage{diagbox}
\usepackage{fullpage}
\usepackage{url}
\usepackage{rotating}
\usepackage{eurosym}
\usepackage{wrapfig}
\usepackage[final]{pdfpages} 
\usepackage{epstopdf} 
\usepackage[a4paper]{geometry}
\usepackage[nottoc]{tocbibind}
\usepackage{placeins}
\usepackage{float}
\usepackage{listings}
\usepackage{color, colortbl}
\usepackage{hyperref}
\usepackage{xcolor}
\usepackage{graphicx, caption, subcaption}
\usepackage[nohyperlinks]{acronym}
\usepackage{sidecap}
\usepackage{algorithm,algpseudocode}
\usepackage{multirow}
\usepackage{booktabs}
\hypersetup{
colorlinks,
citecolor=black,
filecolor=black,
linkcolor=black,
urlcolor=black
} 

\newcommand{\argmin}{\operatornamewithlimits{argmin}}
\newcommand{\sibo}{\textcolor{black}}

\newcommand{\bx}{\textbf{x}}

\newcommand{\by}{\textbf{y}}
\newcommand{\bB}{\textbf{B}}
\newcommand{\bH}{\textbf{H}}
\newcommand{\bR}{\textbf{R}}
\newcommand{\bA}{\textbf{A}}

\newcommand{\bD}{\textbf{D}}

\newcommand{\bY}{\textbf{Y}}
\newcommand{\bC}{\textbf{C}}
\newcommand{\bL}{\textbf{L}}

\newcommand{\bX}{\textbf{X}}

\newcommand{\bI}{\textbf{I}}

\newcommand{\bP}{\textbf{P}}
\newcommand{\bK}{\textbf{K}}

\journal{Journal of computational physics (accepted)}

\begin{document}

\begin{frontmatter}


\title{Efficient deep data assimilation with sparse observations and time-varying sensors}

\author{Sibo Cheng$^{1,2}$,  Che Liu$^{3}$, Yike Guo$^{1,4}$, Rossella Arcucci$^{3}$ }

\address{   \small $^{1}$ Data Science Instituite, Department of Computing, Imperial College London, UK\\
      \small $^{2}$  Leverhulme Centre for Wildfires, Environment, and Society, London, UK\\
        \small $^{3}$ Department of Earth Science \& Engineering, Imperial College London, UK\\
        \small  $^{4}$ Department of Computer Science and Engineering, Hong Kong university of science and technology, China      
}



\begin{abstract}
Variational Data Assimilation (DA) has been broadly used in engineering problems for field reconstruction and prediction by performing a weighted combination of multiple sources of noisy data. 
In recent years, the integration of deep learning (DL) techniques in DA has shown promise in improving the efficiency and accuracy in high-dimensional dynamical systems.
Nevertheless, existing deep DA approaches face difficulties in dealing with unstructured observation data, especially when the placement and number of sensors are dynamic over time.
We introduce a novel variational DA scheme, named Voronoi-tessellation Inverse operator for VariatIonal Data assimilation (VIVID), that incorporates a DL inverse operator into the assimilation objective function. By leveraging the capabilities of the Voronoi-tessellation and convolutional neural networks, VIVID is adept at handling sparse, unstructured, and time-varying sensor data. Furthermore, the incorporation of the DL inverse operator establishes a direct link between observation and state space, leading to a reduction in the number of minimization steps required for DA. Additionally, VIVID can be seamlessly integrated with Proper Orthogonal Decomposition (POD) to develop an end-to-end reduced-order DA scheme, which can further expedite field reconstruction. Numerical experiments in a fluid dynamics system demonstrate that VIVID can significantly outperform existing DA and DL algorithms. The robustness of VIVID is also accessed through the application of various levels of prior error, the utilization of varying numbers of sensors, and the misspecification of error covariance in DA.
\end{abstract}


\end{frontmatter}

\newpage
\section*{Main Notations}
\begin{table*}[ht!]
    \centering
    \begin{tabular}{ p{3.5cm} p{15cm}}
$\bX_t$ & state field in the full space at time $t$\\
$N_x, M_x$ & dimension of the state space\\
$\bx_t$ & flattened state field \\
$\hat{\bx}_t$ & compressed state vector in the reduced space\\
$ \bY_t$ & full observation field at time $t$\\
$N_y, M_y$ & dimension of the observation space\\
$(i_{t,k}, j_{t,k})_{k=1...k^*}$ & sensor positions in observation space \\
$\{y_{t,k}\}_{k=1...k^*} $ & observed values at $(i_{t,k}, j_{t,k})_{k=1...k^*}$ \\
$R_{t,k}$ & the \sibo{Voronoi cell} associated to $y_{t,k}$\\
$ \tilde{\bY}_t$ & tessellated observation field based on $y_{t,k}$ and $R_{t,k}$ \\
$\by_t$ & flattened observation vector\\
${\bx}_{b,t},\hat{\bx}_{b,t}$ & background state in full and reduced space at time $t$\\
${\bx}_{v,t},\hat{\bx}_{v,t}$ & machine-learned state in full and reduced space at time $t$\\
${\bx}_{a,t},\hat{\bx}_{a,t}$ & analysis state in full and reduced space at time $t$\\
$\mathcal{H}_t, \hat{\mathcal{H}}_t$ & transformation operator in full and reduced spaces\\
$\mathcal{X}$ & ensemble of state field snapshots\\
${\bL}_{\mathcal{X},q}$ & POD projection operator with truncation parameter $q$\\
${\bB}_{t}, \hat{\bB}_{t}$ & background error covariance matrices in full and reduced spaces\\
${\bR}_{t}, \hat{\bR}_{t}$ & observation error covariance matrices in full and reduced spaces\\
${\bP}_{t}, \hat{\bP}_{t}$ &  error covariance matrices of the inverse operator in full and reduced spaces\\
$\mathcal{J}_t$ & objective function of variational data assimilation\\
$L$ & correlation scale length in prior error covariances\\
$L^E$ & estimated correlation scale length in prior error covariances\\
    \end{tabular}
\end{table*}

\clearpage

\section{Introduction}
The application of variational \ac{DA} is widespread in engineering problems, serving the purpose of state estimation and/or parameter identification. Variational \ac{DA} is performed through a weighted combination of multiple sources of noisy data~\cite{Carrassi2017}.
 The goal of variational \ac{DA} is to estimate the initial conditions of a system such that its subsequent evolution matches with real-world observations. The \ac{DA} technique involves solving an optimization problem to find the best initial conditions or states and then integrating the system equations of motion over time to make predictions. \ac{DA} has been widely applied in ocean engineering~\cite{elisseeff2002ocean,Carrassi2017}, \ac{NWP}~\cite{lorenc2015comparison}, nuclear engineering~\cite{gong2022data} and hydrology~\cite{liu2012advancing, cheng2021error}. 
 However, traditional \ac{DA}  algorithms are encumbered by the computational demands arising from the numerous iterations necessary to solve the intricate optimization problem, particularly in cases where the observation data is unstructured and the placement of sensors is dynamic over time~\cite{apte2008data,ghil1991data}.
 
 In recent years, the use of machine learning in conjunction with \ac{DA} has shown promise in further improving the efficiency and accuracy of predictions in high-dimensional dynamical systems~\cite{cheng2023machine,farchi2021comparison, Cheng2022JCP,Cheng2022JSC,tang2020deep,pawar2020long,LIU202246}. Among these approaches, many of them aim to \sibo{surrogate directly}  some key elements in \ac{DA} such as the forward operator~\cite{LIU202246, Cheng2022JCP, pawar2020long,casas2020}, the transformation function~\cite{wang2022deep}, the \ac{ROM}~\cite{Cheng2022JSC,peyron2021latent} and the optimization process~\cite{gottwald2021supervised} using \ac{ML} techniques. \sibo{Recent research}~\cite {fablet2022multimodal, filoche2022variational} also aims to build end-to-end \ac{DL} frameworks to emulate the entire \ac{DA} procedure. 
In essence, the use of physics-based models in \ac{DA} is being replaced, to some extent, by data-driven operators.
  Although machine learning models are powerful and efficient, physical models still provide benefits such as generalizability, interpretability, and consistency in established physical principles~\cite{geer2021learning}. Additionally, physical models offer the advantage of coupling processes of different spatial and temporal scales, which is vital for complex forecasting applications~\cite{stauffer1994multiscale}. Compared to pure data-driven \ac{DA} methods, a more promising path might be to enhance physics-driven \ac{DA} schemes with \ac{ML}.

Given these circumstances, significant efforts have been directed towards enhancing the efficiency of physics-driven \ac{DA} through the utilization of \ac{ML} techniques~\cite{chattopadhyay2023deep,ouala2018neural,buizza2022data}, rather than solely relying on \ac{ML} to replace traditional \ac{DA} methods. In particular, the recent work of~\cite{pmlr-v139-frerix21a} introduces the learned observation operator in the variational assimilation scheme, enhancing and facilitating the optimization of the \ac{DA} objective function. More precisely, a \ac{DL} model is built to map the observation space to the state space.  
The assimilation scheme proposed by~\cite{pmlr-v139-frerix21a} then consists of a two-stage minimization procedure with the machine-learned inverse operator and the full transformation function, respectively. The former acts as an initialization step of the latter, which is a conventional variational assimilation algorithm. In other words,~\cite{pmlr-v139-frerix21a} doesn't modify the objective function in \ac{DA} and the \ac{ML} inverse operator is only used to provide an advanced initialization point in the subsequent optimization. In addition, based on a traditional \ac{CNN} structure, their method can only handle structured observations (e.g., square grids) with a fixed number of sensors (if not retrained) which limits its application in real-world problems where incomplete/sparse observations and time-varying sensors are prevalent~\cite{barker2004three, elbern2001ozone}. 

Geometric deep learning methods, namely \ac{GNN}, have been used to deal with unstructured data~\cite{zhou2020fully,shi2022gnn}. However, it is widely noticed that training high-dimensional \ac{GNN} can be computationally expensive and time-consuming~\cite{wu2020comprehensive, zhou2020graph}. The recent work of~\cite{fukami2021global} makes use of Voronoi tessellation~\cite{watson1981computing} to address the bottleneck of sparse observations and various numbers of sensors in \ac{CNN}-based field reconstruction.
More precisely, a Voronoi tessellation is used to partition the input data into distinct regions based on the proximity of placed sensors. The resulting tessellation can be used to generate a set of Voronoi cells, where each cell represents a distinct region of the input space. This tessellation is then used as input to train a \ac{CNN} model to reconstruct the whole underlying physical field. Such method, known as \ac{VCNN}, demonstrates strong performance in both synthetic \ac{CFD} data and real geophysical fields~\cite{fukami2021global}, compared to traditional field interpolation approaches such as Kriging~\cite{kleijnen2009kriging}. However, it is worth mentioning that mapping the sparse observation space to the whole physical field is often ill-defined. As a consequence, the prediction can be biased by the training dataset. Integrating a prior estimation, as performed in \ac{DA}, can constrain the full physical space, thus addressing the issue of an ill-defined system. To the best of the author's knowledge, none \ac{DA} or Bayesian approaches with \ac{VCNN} has been proposed in the existing literature.  

In this paper, we introduce a novel \ac{DL}-assisted \ac{DA} approach, named \ac{VIVID}, that couples a \ac{VCNN}-based inverse operator in the \ac{DA}. Unlike~\cite{pmlr-v139-frerix21a}, \ac{VIVID} is an end-to-end \ac{DA} scheme that minimizes jointly the background mismatch in the state space, the inverse mismatch in the state space and the observation mismatch in the observation space. Similar to \ac{DA}, the weight of different components in the objective function is determined by the estimated error covariance matrices. In fact, conventional \ac{DA} can provide accurate local correction where the observations are dense, while \ac{VCNN} is capable of delivering a global field prediction based on the knowledge extracted from historical data. By definition, the proposed \ac{VIVID} incorporates the strength of both methods. In addition, by including a prior state-observation mapping provided by \ac{VCNN}, the optimization of the \ac{DA} objective function can be speeded-up, resulting in potentially fewer minimization iterations. To reduce the computational burden of \ac{DA} for high-dimensional systems, much effort is given to performing \ac{DA} in reduced order spaces~\cite{Cheng2022JCP,xiao2018parameterised,LIU202246}. In this paper, we incorporate a projection-based \ac{ROM}, namely \ac{POD}, with the novel \ac{VIVID} approach. In VIVID-ROM, instead of performing \ac{POD} on top of the inverse operator, we build an end-to-end neural network that directly maps the tessellated observation space to the reduced state space with a similar structure as a vision encoder~\cite{wen2018neural}. VIVID-ROM, as a variant of \ac{VIVID}, clearly brings more insight into applying the developed approach in mainstream \ac{DA} frameworks where \ac{ROM} is often required. 

As mentioned in~\cite{pmlr-v139-frerix21a}, in many applications (e.g., \ac{NWP}, geophysics), ample training data from historical observations can be found to train \ac{VCNN} and \ac{VIVID}.
As a proof of concept, we perform extensive numerical experiments in this study on a shallow water \ac{CFD} model~\cite{Venant1871}. The proposed \ac{VIVID} is compared against the conventional variational \ac{DA} and \ac{VCNN} in terms of both reconstruction accuracy and computational efficiency. The reconstruction is evaluated using \ac{R-RMSE} and \ac{SSIM} where the former computes the pixel-wise reconstruction error and the latter measures the global similarity between reconstructed fields and the ground truth. In addition, we extensively assess the robustness of different approaches regarding background error, observation error, varying number of sensors and \sibo{error covariance misspecification}. The last one significantly impacts assimilation performance, raising broad research interests in the past two decades~\cite{tandeo2020review,cheng2022observation}.

 In summary, we make the following main contributions in this study,
 \begin{itemize}
     \item We introduce a novel \ac{DL}-assisted \ac{DA} approach \ac{VIVID} and its reduced order variant \ac{VIVID}-\ac{ROM}, which is capable of assimilating data from sparse, position-varying and number-varying sensors that current Deep \ac{DA} approaches (e.g.,~\cite{peyron2021latent,Cheng2022JCP,pawar2020long,wang2022deep}) can not handle.
     \item Numerical experiments on a two-dimensional \ac{CFD} test case show that the proposed \ac{VIVID} substantially outperforms the variational \ac{DA} and the state-of-the-art \ac{DL} field reconstruction method \ac{VCNN} by reducing the R-RMSE of around 50\%.  \ac{VIVID} also requires around $70\%$ less computational time compared to conventional \ac{DA}. 
     \item The robustness of the proposed approach, regarding different noise levels and \ac{DA} assumptions, is extensively studied in this paper. Within all numerical experiments implemented, \ac{VIVID} achieves an outstanding performance in comparison with variational \ac{DA} and \ac{VCNN}.
 \end{itemize}
 
 The rest of this paper is organized as follows. Section~\ref{sec:DA} reminds the formulation of variational \ac{DA} and the minimization loops of its objective function. Section~\ref{sec:methodology} introduces the methodology of \ac{VIVID} and its reduced order variant \ac{VIVID}-\ac{ROM}. Numerical experiments on a shallow water model with various conditions and assumptions are presented in Section~\ref{sec:experiments}. We close the paper with a discussion in Section~\ref{sec:discuss}.
 
\section{Background: Variational data assimilation}
\label{sec:DA}

The goal of \ac{DA} is to enhance the identification/reconstruction of some physical fields $\bX_t \in \mathbb{R}^{\{N_x, M_x\}}$ at a given time step $t$. Let us denote $\bx_t$, the flatten vector associated to $\bX_t$, i.e., $\bx_t \in \mathbb{R}^{\{N_x \times M_x,1\}}$.  \ac{DA} makes use of two types of information: a prior forecast of the current state vector $\bx_t$, also known as the background state $\textbf{x}_{b,t}$, and an observation vector represented by $\textbf{y}_t$.
The theoretical value of the state variable at time $t$ is denoted by $\textbf{x}_{\textrm{true},t}$, also known as the true state. 
Variational \ac{DA} seeks the optimal balance between $\textbf{x}_{b,t}$ and $\textbf{y}_t$ by minimizing the cost function $J$ that is defined as:
\begin{align}
    J_t(\textbf{x})&=\frac{1}{2}(\textbf{x}-\textbf{x}_{b,t})^T\textbf{B}_t^{-1}(\textbf{x}-\textbf{x}_{b,t}) + \frac{1}{2}(\textbf{y}_t-\mathcal{H}_t(\textbf{x}))^T \textbf{R}_t^{-1} (\textbf{y}_t-\mathcal{H}_t(\textbf{x}_t)) \label{eq_3dvar}\\
   &=\frac{1}{2}\vert \vert\textbf{x}-\textbf{x}_{b,t}\vert \vert^2_{\textbf{B}_t^{-1}}+\frac{1}{2}\vert \vert\textbf{y}_t-\mathcal{H}_t(\textbf{x})\vert \vert^2_{\textbf{R}_t^{-1}} \notag
\end{align}
 where  $\mathcal{H}_t$ is the state-observation mapping function \sibo{that links the flattened state vector to the flattened observation vector.} $(\cdot)^T$ in Equation ~\eqref{eq_3dvar}  is the transpose operator.  $\textbf{B}_t$ and $\textbf{R}_t$ denote the error covariance matrices in relation to  $\textbf{x}_{b,t}$ and  $\textbf{y}_t$, that is,
 \begin{align}
     \textbf{B}_t = \textrm{Cov}(\epsilon_{b,t}, \epsilon_{b,t}), \quad
     \textbf{R}_t = \textrm{Cov}(\epsilon_{y,t}, \epsilon_{y,t}),
 \end{align}
 where
  \begin{align}
     \epsilon_{b,t} = \textbf{x}_{b,t} - \textbf{x}_{\textrm{true},t}, \quad
     \epsilon_{y,t} = \mathcal{H}_t(\textbf{x}_{\textrm{true},t})-\textbf{y}_t.
 \end{align}
 Prior errors $\epsilon_b, \epsilon_y$ are often \sibo{assumed} to be \sibo{centred Gaussian} thus they can be fully characterised by $\textbf{B}_t$ and $\textbf{R}_t$,
 \begin{align}
     \epsilon_{b,t} \sim \mathcal{N} (0, \textbf{B}_t), \quad
     \epsilon_{y,t} \sim \mathcal{N} (0, \textbf{R}_t).
 \end{align}
Equation~\eqref{eq_3dvar} is known as the three-dimensional variational (3D-Var) formulation, yielding a general objective function of variational \ac{DA}. 
The point of minimum in Equation~\eqref{eq_3dvar} is the analysis state $\bx_{a,t}$, 
  \begin{align}
    \bx_{a,t} = \underset{\bx}{\argmin} \Big(J_t(\textbf{x})\Big) \label{eq:argmin}.
 \end{align}
 
  In the case where $\mathcal{H}_t$ can be approximated by a linear function $\bH_t$, the optimization of Equation~\eqref{eq_3dvar} can be solved by \ac{BLUE}~\cite{Carrassi2017}, 
 \begin{equation}
	\bx_{a,t}=\bx_{b,t}+\bK_t(\by_t-\bH_t \bx_{b,t}) \label{eq:BLUE_1},
\end{equation} 
where the Kalman gain matrix $\bK_t$ is defined as
\begin{equation}
	\bK_t=\bB_t \bH_t^T (\bH_t \bB_t \bH_t^T+\bR_t)^{-1}. 
	\label{eq:Kgain_BLUE}
\end{equation}

Following Equation~\eqref{eq:BLUE_1}, we can also obtain an output error covariance estimation $\bA_t$,
\begin{align}
    \bA_t=(\textbf{I}-\bK_t\bB_t\bH_t)\textbf{B}_t. \label{eq:assumed_A}
\end{align}

The trace of the analysis covariance matrix, i.e., $Tr(\bA_t)$, representing the total variance of posterior error, is often used as an indicator to evaluate the assimilation accuracy~\cite{Carrassi2017,Eyre2013}.

  When $\mathcal{H}_t$ is non-linear, approximate iterative methods~\cite{lawless2005approximate} have been widely used to solve variational data assimilation. To do so, one has to compute the gradient $\nabla J(\bx)$, which can be approximated by
 \begin{align}
     \nabla J(\bx) \approx 2 \bB_t^{-1}(\bx-\bx_{b,t}) - 2 \bH^T \bR_t^{-1} (\by_t-\mathcal{H}_t(\bx)). \label{eq:gradient}
 \end{align}
 In equation~\eqref{eq:gradient}, $\bH$ is obtained via a local linearization \sibo{in the neighborhood of the current vector $\bx$}. The minimization of 3D-Var is often performed via quasi-Newton methods, including for instance, BFGS approaches~\cite{Fulton2000}, where each iteration can be written as:
  \begin{align}
     \bx_{k+1} = \bx_{k} - L_\textrm{3D-Var} \big[ \textrm{Hess}(J) (\bx_{k})\big]^{-1} \nabla J(\bx_k)
 \end{align}
Here $k$ is the current iteration, and $L_\textrm{3D-Var}>0$ is the learning rate of the descent algorithm, and
\begin{align}
\textrm{Hess}\Big(J(\bx = [x_0,...,x_{n-1}])\Big)_{i,j} = \frac{\partial^2 J}{\partial x_i \partial x_j}
\end{align}
is the Hessian matrix related to the cost function $J$. \sibo{The process of the iterative minimization algorithm is summarised in Algorithm~\ref{algo:1}, where the approximation of the Hessian matrix is derived from the analytical formulation by solving only first-order sensitivities \cite{shi2018approximate}. }

\begin{algorithm}[]

\caption{Iterative minization of 3D-Var cost function via quasi-Newton methods }
\begin{algorithmic}[1]
\State Inputs: $\bx_{b,t}, \by_t, \bB_t, \bR_t, \mathcal{H}_t$
\State parameters: $k_\textrm{max}$ , $\epsilon$ 
\State $\bx_0 = \bx_b,$ $ k = 0$\\

\While{$k<k_\textrm{max}$ \textrm{and} $\vert \vert \nabla J_t(\bx_k)\vert \vert  > \epsilon$ }
        \State $J_t(\textbf{x}_k) = \frac{1}{2}\vert \vert\textbf{x}_k-\textbf{x}_{b,t}\vert \vert^2_{\textbf{B}_t^{-1}}+\frac{1}{2}\vert \vert\textbf{y}_t-\mathcal{H}_t(\textbf{x}_k)\vert \vert^2_{\textbf{R}_t^{-1}}$ 
        \State linearize  $\bH_t \approx \mathcal{H}_t$ in the neighbourhood of $\bx_k$
        \State $\nabla J_t(\bx_k) \approx 2 \bB_t^{-1}(\bx_k-\bx_{b,t}) - 2 \bH^T \bR_t^{-1} (\by_t-\mathcal{H}_t(\bx_k))$
        \State compute \sibo{the approximated Hessian matrix} $\textrm{Hess}\big(J_t(\bx_k)\big)$
        \State $ \bx_{k+1} = \bx_{k} - L_\textrm{3D-Var} \big[ \textrm{Hess}(J) \bx_{k}\big]^{-1} \nabla J_t(\bx_k)$
        \State \textit{k} = \textit{k}+1
\EndWhile
\end{algorithmic}
output: $\bx_k$

\label{algo:1}
\end{algorithm}

\section{Methodology}
\label{sec:methodology}
In this section, we describe the methodology of the proposed deep data assimilation scheme \ac{VIVID}, including a machine learning inverse operator for unstructured observations and the \ac{DL}-assisted variational \ac{DA} formulation.
\subsection{Voronoi tessellation for inverse operator}
In this study, with a similar idea as~\cite{fukami2021global}, we build an inverse operator to map sparse and unstructured observations to the state space. 
\sibo{We consider an observation field $\bY_t \in \mathbb{R}^{N_y  \times M_y}$ as a set of observable points (sensors) at a given time $t$}, located at $\{(i_{t,k},j_{t,k})\} $ where
\begin{align}
\{i_{t,k},j_{t,k}\} \in [1,...,N_y] \times [1,...,M_y] \quad \textrm{and}\quad \textrm{$k*$ is the number of sensors}. 
\end{align}

$\{y_{t,k}\}_{k=1...k^*}$ are the observed values. A Voronoi cell $R_{t,k}$ associated to the observation $\{y_{t,k}, i_{t,k},j_{t,k}\}$ can be defined as
\begin{align}
    R_{t,k} = \big\{ \{i_y,j_y\}  \hspace{2mm} | \hspace{2mm} d((i_y,j_y), (i_{t,k},j_{t,k})) \leq d((i_y,j_y), (i_{t,q},j_{t,q})), \hspace{2mm} \forall \hspace{2mm} 1 \leq q \leq k^* \hspace{2mm} \textrm{and} \hspace{2mm} q \neq k \big\}.
\end{align}
\sibo{Here, $d(\cdot)$ is the Euclidean distance}, i.e.,
\begin{align}
    d((i_y,j_y), (i_{t,k},j_{t,k})) = \sqrt{(i_y-i_{t,k})^2+(j_y-j_{t,k})^2},
\end{align}
where $ \{i_y,j_y\} \in [1,...,N_y] \times [1,...,M_y]$. Therefore, the observation space can be split into several Voronoi cells regardless of the number of sensors, that is,
\begin{align}
    \{(i_y,j_y)\} = \bigcup_{k=1}^{k*} R_{t,k} \quad \textrm{and} \quad R_{t,k} \cap R_{t,q} = \emptyset \quad (\forall  \quad k \neq q).
\end{align}
 A tessellated observation $\tilde{\bY}_t = \{ \tilde{y}_{t,i_y,j_y}\} \in \mathbb{R}^{N_y  \times M_y}$ in the full observation space can be obtained by
\begin{align}
    \tilde{y}_{t,i_y,j_y} = y_{t,k} \quad \textrm{if} \quad (i_y,j_y) \in R_{t,k}.
\end{align}

The tessellation $\tilde{\bY}_t$  is then used as input for training a deep learning model to map the unobservable state field $\bX_t$. To do so, a \ac{CNN} is employed, i.e.,
\begin{align}
    \tilde{\bY}_t \xrightarrow{\hspace*{0.6cm} \text{CNN} \hspace*{0.6cm}} \bX_t. \label{eq:CNN}
\end{align}

\begin{figure*}[h!]
\centering
\includegraphics[width=0.7\textwidth]{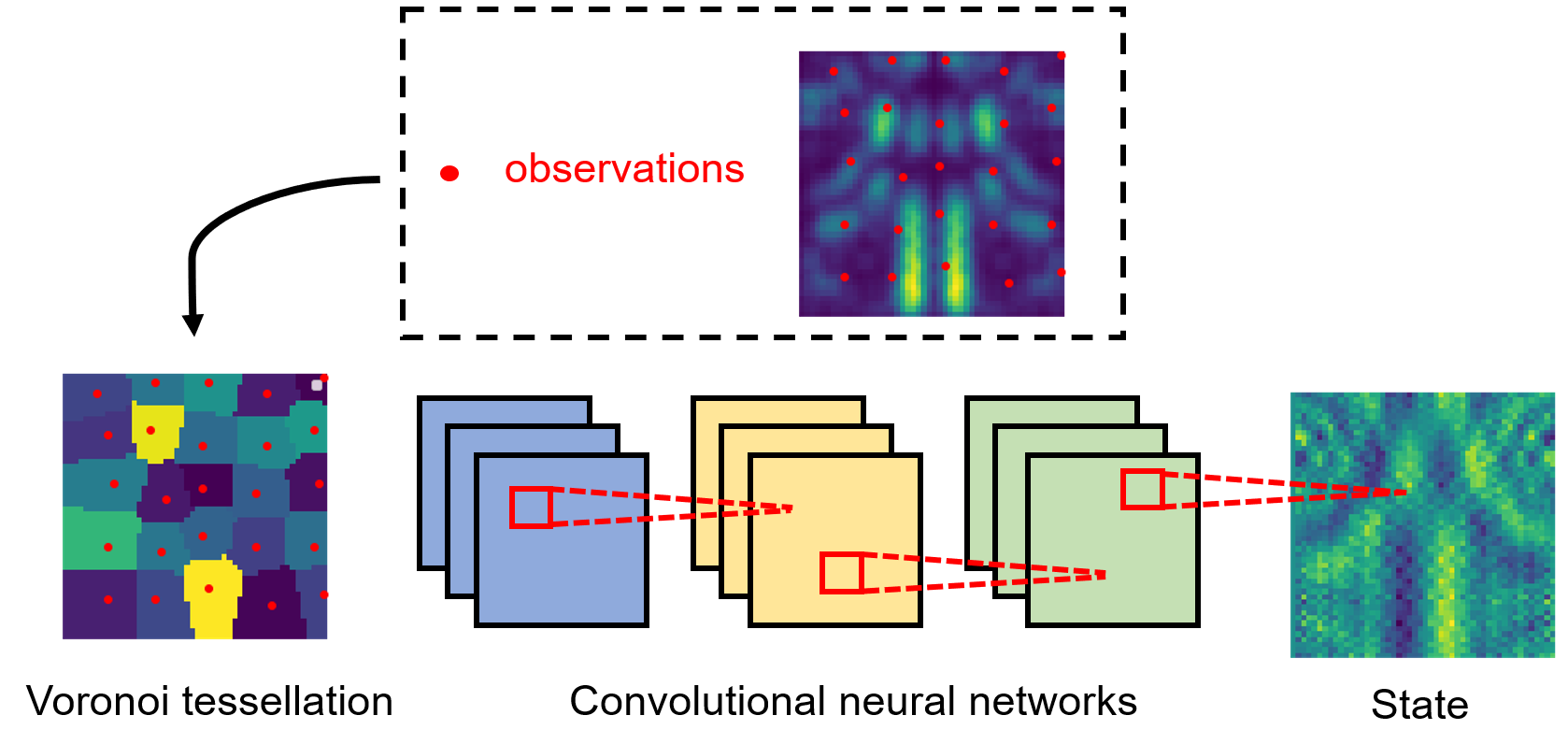}
\caption{Flowchart of the Voronoi tessellation-assisted CNN to map the state field using sparse observations}
\label{fig:CNN}
\end{figure*}

The \ac{CNN} implemented in this study consists of 7 convolution layers with ReLu activation function~\cite{wang2019learning} and 48 channels~\cite{fukami2021global}. The large number of channels ensure the representation capacity of the \ac{NN} and, more importantly, improve the robustness to variations regarding different sensor placement. A cropping or upsampling layer is also added to reshape the two-dimensional field from $(N_y,M_y)$ to $(N_x,M_x)$, if different. The workflow of Voronoi tessellation-assisted observation-state mapping is illustrated in Figure~\ref{fig:CNN} and
the exact \ac{CNN} structure is shown in Table~\ref{table: CAE_structure}.

\begin{table}[h!]
\centering
\caption{NN structure of the \ac{CNN} with $\tilde{\bY}_t$ as input and ${\bX}_t$ as output}
\begin{tabular}{ccc} \toprule
    {\textbf{Layer (type)}} 
    & {\textbf{Output Shape}} 
    & {\textbf{Activation}} \\ \midrule
     & &  \\
    {Input}  
    & {$(N_y, M_y,1)$}
    & {}\\
    {\textrm{Conv 2D} $(8\times8)$}  
    & {$(N_y, M_y,48)$}
    & {ReLu}\\
    {\textrm{Conv 2D} $(8\times8)$}  
    & {$(N_y, M_y,48)$}
    & {ReLu}\\
    {\textrm{Conv 2D} $(8\times8)$}  
    & {$(N_y, M_y,48)$}
    & {ReLu}\\
    {\textrm{Conv 2D} $(8\times8)$}  
    & {$(N_y, M_y,48)$}
    & {ReLu}\\
    {\textrm{Conv 2D} $(8\times8)$}  
    & {$(N_y, M_y,48)$}
    & {ReLu}\\
    {\textrm{Conv 2D} $(8\times8)$}  
    & {$(N_y, M_y,48)$}
    & {ReLu}\\  
    {\textrm{Cropping 2D}}  
    & {$(N_x, M_x,48)$}
    & \\ 
    {\textrm{Conv 2D}}  
    & {$(N_x, M_x,1)$}
    & {ReLu}\\ 
    \bottomrule
\end{tabular}
\label{table: CAE_structure}
\end{table}

It is worth mentioning that despite the dimension of $\tilde{\bY}_t$ equals to $\{N_y, M_y\}$, by construction, its degree of freedom remains as the number of sensors $k$. As a consequence, Equation~\eqref{eq:CNN} represents an under-determined system when $k<N_x \times M_y$. This is most likely since the observations in \ac{DA} are often assumed sparse and incomplete~\cite{Carrassi2017}. Therefore, performing field reconstruction using solely \ac{VCNN} without \ac{DA} can lead to ill-defined problems.

\subsection{Voronoi-tessellation Inverse operator for VariatIonal
Data assimilation (VIVID)}
\begin{figure*}[h!]
\centering
\includegraphics[width=0.9\textwidth]{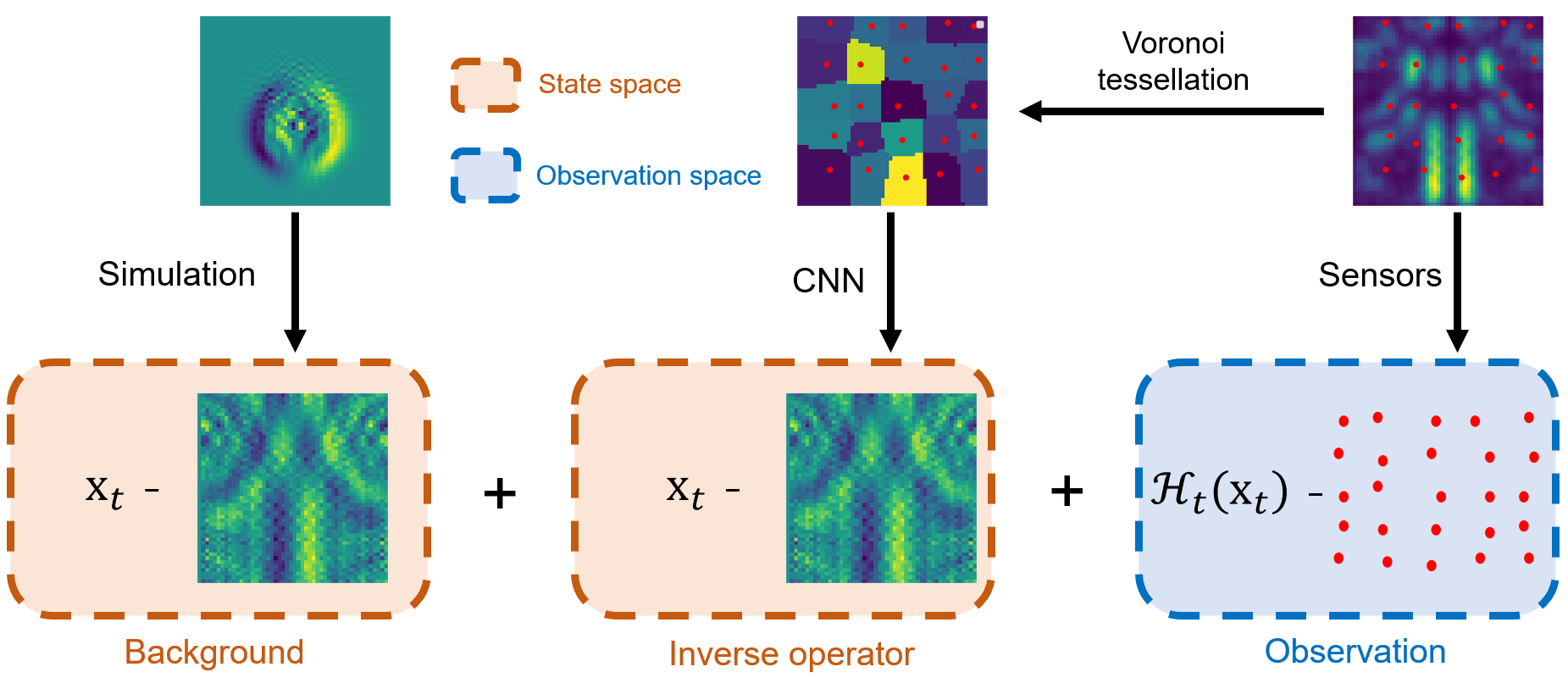}
\caption{Flowchart of the proposed deep DA approach}
\label{fig:VIVID}
\end{figure*}

In the proposed \ac{DA} approach of this paper, the \sibo{machine-learned observation-state operator} plays a pivotal role in enhancing the \ac{DA} accuracy and accelerating the minimization convergence. Denoting $\bx_{v,t} = \textrm{CNN}(\tilde{\bY}_t)$ as the flattened output of the \ac{CNN} (Table~\ref{table: CAE_structure}) applied on $\tilde{\bY}_t$, the objective function of the proposed variational \ac{DA} reads
\begin{align}
    J_t(\textbf{x})&=\frac{1}{2}\vert \vert\textbf{x}-\textbf{x}_{b,t}\vert \vert^2_{\textbf{B}_t^{-1}}+\frac{1}{2}\vert \vert\textbf{x}-\bx_{v,t}\vert \vert^2_{\textbf{P}_t^{-1}}+\frac{1}{2}\vert \vert\textbf{y}_t-\mathcal{H}_t(\textbf{x})\vert \vert^2_{\textbf{R}_t^{-1}}, \label{eq:DA_op}
\end{align}
where $\bP_t$ denotes the error covariance matrix of the learned operator. The minimization of Equation~\eqref{eq:DA_op} follows the same process as the traditional variational \ac{DA}. The proposed \ac{DA} algorithm scheme can be summarized in Figure~\ref{fig:VIVID} and algorithm~\ref{algo:2}.

\begin{algorithm}[]

\caption{Formulation and minization of the proposed variational DA }
\begin{algorithmic}[1]
\State Inputs: $\bx_{b,t}, \by_t, \bB_t, \bR_t, \bP_t,\mathcal{H}_t$
\State parameters: $k_\textrm{max}$ , $\epsilon$ 
\State $\bx_0 = \bx_b,$ $ k = 0$\\
$\tilde{\bY}_t \xleftarrow{\hspace*{0.6cm} \text{Voronoi} \hspace*{0.6cm}} \by_t$\\
compute $\bx_{v,t} = \textrm{CNN}(\tilde{\bY}_t)$\\

\While{$k<k_\textrm{max}$ \textrm{and} $\vert \vert \nabla J_t(\bx_k)\vert \vert  > \epsilon$ }
        \State $J_t(\textbf{x}_k) = \frac{1}{2}\vert \vert\textbf{x}_k-\textbf{x}_{b,t}\vert \vert^2_{\textbf{B}_t^{-1}}+\frac{1}{2}\vert \vert\textbf{x}_k-\bx_{v,t}\vert \vert^2_{\textbf{P}_t^{-1}}+\frac{1}{2}\vert \vert\textbf{y}_t-\mathcal{H}_t(\textbf{x}_k)\vert \vert^2_{\textbf{R}_t^{-1}}$ 
        \State linearize  $\bH_t \approx \mathcal{H}_t$ in the neighbourhood of $\bx_k$
        \State $\nabla J_t(\bx_k) \approx 2 \bB_t^{-1}(\bx_k-\bx_{b,t}) + 2 \bP_t^{-1}(\bx_k-\bx_{v,t}) - 2 \bH_t^T \bR_t^{-1} (\by_t-\mathcal{H}_t(\bx_k))$
        \State compute \sibo{the approximated Hessian matrix} $\textrm{Hess}\big(J_t(\bx_k)\big)$
        \State $ \bx_{k+1} = \bx_{k} - L_\textrm{3D-Var} \big[ \textrm{Hess}(J) \bx_{k}\big]^{-1} \nabla J_t(\bx_k)$
        \State \textit{k} = \textit{k}+1
\EndWhile
\end{algorithmic}
output: $\bx_k$

\label{algo:2}
\end{algorithm}

As for the specification of $\bP_t$, it can be empirically estimated using an independent validation dataset which is different from the \sibo{training dataset}, that is,
\begin{align}
    {\textbf{P}}_t\approx\frac{1}{n_{\text{val}}-1}\sum_{\text{val dataset}}({\bx}_{t}-\bx_{v,t})\left({\bx}_{t}-\bx_{v,t}\right)^T. 
    \label{eq:ensemble_P}
\end{align}
To regularize the estimated ${\textbf{P}}_t$ matrix, covariance localization is implemented in this study, thanks to the Gaspari-Cohn localization function~\cite{Gaspari1998} defined as:

\begin{align}
  G(\rho) =
    \begin{cases}
      \text{if 0 $\leq$ $\rho$ $<$ 1: } &1-\frac{5}{3}\rho^2+\frac{5}{8}\rho^3 +\frac{1}{2}\rho^4-\frac{1}{4}\rho^5\\
      \text{if 1 $\leq$ $\rho$ $<$ 2: } &
      4-5\rho+\frac{5}{3}\rho^2+\frac{5}{8}\rho^3- \frac{1}{2}\rho^4+\frac{1}{12}\rho^5-\frac{2}{3\rho}
      \\\
      \text{if $\rho$ $\geq$ 2: } &0
    \end{cases}  \quad \textrm{with} \quad \rho = \frac{r}{L}. \label{eq:Gaspari}
\end{align}
Here, $r$ denotes the spatial distance of two points in the state space and $L$ is an empirically defined correlation scale length. The localized covariance matrix can then be obtained,
\sibo{\begin{align}
    \forall {{P}}_{t}^{a,b} \in {\textbf{P}}_t, \quad {P}_{t}^{a,b} \longleftarrow {P}_{t}^{a,b}\cdot G(\rho), \quad \text{with} \quad \omega = \frac{d(a,b)}{L}, \quad \{a,b\} \in [1,...,N_x \times M_x], \label{eq:Gasp}
\end{align}}
where \sibo{$\omega(\cdot)$} denotes a distance metric in the state space. Here we remind that ${\textbf{P}}_t \in \mathbb{R}^{(N_x  \times M_x)^2}$ is the error covariance matrix of flattened state vectors. Thus, the locations $a,b$ are with two-dimensional indices in the state space.

\subsection{Posterior error analysis for a linear case}
Considering a new observation vector $\bar{\by}_t$ as the concatenation of ${\bx}_{v,t}$ and $\by_t$. Equation~\eqref{eq:DA_op} can be re-written as:
\begin{align}
    J_t(\textbf{x})=\frac{1}{2}\vert \vert\textbf{x}-\textbf{x}_{b,t}\vert \vert^2_{\textbf{B}_t^{-1}}+\frac{1}{2}\vert \vert\bar{\textbf{y}}_t-\bar{\mathcal{H}}_t(\textbf{x})\vert \vert^2_{\bar{\textbf{R}}_t^{-1}} \label{eq:DA_trans},
\end{align}
where
\begin{align}
\bar{\by}_t = \begin{pmatrix}
        \bx_{v,t}  \\
		{\textbf{y}}_t  
    \end{pmatrix}, \quad 
\bar{\bH}_t = \begin{pmatrix}
        \bI  \\
		{\bH}_t 
    \end{pmatrix},\quad
    \bar{\bR}_t = \begin{pmatrix}
        \bP_t& 0  \\
		0& {\bR}_t 
    \end{pmatrix}.
\end{align}
Similar to Equations~\eqref{eq:assumed_A} and~\eqref{eq:BLUE_1}, when $\mathcal{H}_t$ can be approximated by a linear operator $\mathcal{H}_t \approx \bH_t$, the minimization of Equation~\eqref{eq:DA_trans} can be performed through the \ac{BLUE} formulation,

 \begin{align}
	\bx_{a,t}&=\bx_{b,t}+\bar{\bK}_t(\bar{\by}_t-\bar{\bH}_t \bx_{b,t}) \label{eq:BLUE_2}, \\
 \bA_t&=(\textbf{I}-\bar{\bK}_t\bar{\bH}_t)\textbf{B}_t. \label{eq:assumed_A2}
\end{align} 
with the modified Kalman gain matrix,
\begin{equation}
	\bar{\bK}_t=\bB_t \bar{\bH}_t^T (\bar{\bH}_t \bB_t \bar{\bH}_t^T+\bar{\bR}_t)^{-1} = \big(\bB_t,\bB_t \bH_t\big) \cdot \begin{pmatrix}
        \bB_t + \bP_t & \bB_t \bH_t^T  \\
		\bH_t \bB_t & \bH \bB_t \bH_t^T + \textbf{R}_t
    \end{pmatrix}^{-1}. 
	\label{eq:Kgain_BLUE2}
\end{equation}

 For the simplicity of illustration, we show the evolution of $Tr(\bA_t)$ in function of $\bB_t$, $\bP_t$ and $\bR_t$ in a simple scalar case as being done in~\cite{Eyre2013}. In the particular case of a one-dimensional problem, $\{\bA_t(=Tr(\bA_t)), \bB_t, \bR_t, \bP_t, \bH_t\} \in \mathbb{R}^5$. Therefore Equation~\eqref{eq:assumed_A} and~\eqref{eq:assumed_A2} can be simplified to:
\begin{align}
\bA^{\textrm{DA}}_t &= \frac{\bB_t \bR_t}{\bB_t \bH^2_t + \bR_t},\\
    \bA^{\textrm{VIVID}}_t &=\bB_t\Bigg(\frac{\bB_t^2 \bH^2_t}{\bB_t \bH^2_t \bP_t+\bB_t \bR_t+\bP_t \bR_t}-\frac{\bB_t\left(\bB_t \bH^2_t+\bR_t\right)}{\bB_t \bH^2_t \bP_t+\bB_t \bR_t+\bP_t \bR_t} \notag\\
    &-\bH_t\left(-\frac{\bB_t^2 \bH_t}{\bB_t \bH^2_t \bP_t+\bB_t \bR_t+\bP_t \bR_t}+\frac{\bB_t \bH_t(\bB_t+\bP_t)}{\bB_t \bH^2_t \bP_t+\bB_t \bR_t+\bP_t \bR_t}\right)+1\Bigg). 
\end{align}

\begin{figure}[h!]
\centering
\includegraphics[width = \textwidth]{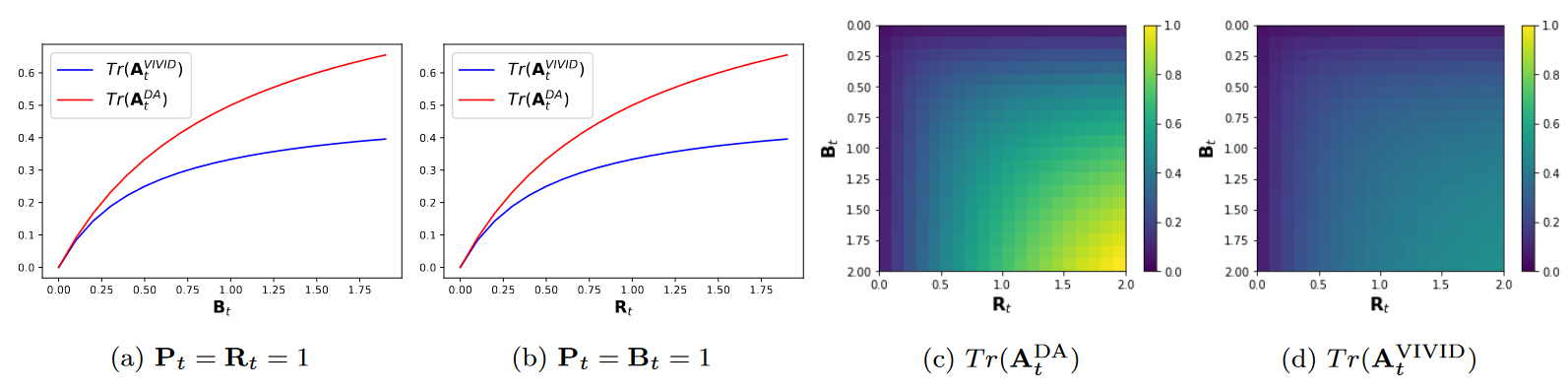}
\caption{The evolution of $Tr(\mathbf{A}^{\textrm{DA}}_t)$ and $Tr(\mathbf{A}^{\textrm{VIVID}}_t)$ regarding $\bB_t$ and $\bR_t$}
\label{fig:scalar}
\end{figure}

We are interested in the sensitivity of $Tr(\bA^{\textrm{DA}}_t)$ and $Tr(\bA^{\textrm{VIVID}}_t)$ with respect to $\bB_t$, $\bR_t$ and $\bP_t$. Without loss of generality, we suppose $\bH_t = \bP_t = 1$ in the following illustrations while varying the values of $\bB_t$ and $\bR_t$. Figure~\ref{fig:scalar} (a) (resp. (b)) illustrates the evolution of $Tr(\mathbf{A}^{\textrm{DA}}_t)$ and $Tr(\mathbf{A}^{\textrm{VIVID}}_t)$ while varying $\bB_t$ (resp. $\bR_t$) from 0 to 2. Since the transformation matrix is set to be $\bH_t = 1$, as expected, symmetric curves can be found in Figure~\ref{fig:scalar} (a) and (b). The posterior error variance can be significantly reduced thanks to \ac{VIVID}, indicating a more accurate assimilation. Figure~\ref{fig:scalar} (c) (resp. (d)) illustrates the value of $Tr(\mathbf{A}^{\textrm{DA}}_t)$ (resp. $Tr(\mathbf{A}^{\textrm{VIVID}}_t)$) while varying simultaneously $\bB_t$ and $\bR_t$. Consistent with Figure~\ref{fig:scalar} (a) and (b), we find that the posterior error variance can be substantially reduced, especially when $\bB_t$ and $\bR_t$ are large. The results of this simple scalar problem demonstrate the strength when introducing another prior estimation, independent of the background state. A detailed numerical comparison of the proposed \ac{VIVID} approach against conventional \ac{DA} and \ac{VCNN}, is performed with a CFD experiment in Section \ref{sec:experiments}.

\subsection{VIVID with reduced order modelling}
Performing \ac{DA} in the full physical space, even with the help of the machine learning inverse operator, can be computationally expensive and time consuming due to the large dimension of the state space. Here, we explain how the proposed approach can be coupled with a \ac{ROM} using \ac{POD} to further improve its efficiency.

Considering a set of $n_\textrm{state}$ state snapshots, from one or several simulations/predictions, are represented by a matrix $\mathcal{X} \in \mathbb{R}^{(\{N_x \times M_x) \times n_\textrm{state}\}}$ where each column of $\mathcal{X}$ represents a flattened state at a given time step, that is,
\begin{align}
    \mathcal{X} = \big[ \bx_0 \big| \bx_1 \big| ...\big| \bx_{n_\textrm{state}-1} \big].
\end{align}

The empirical covariance $\bC_{\mathcal{X}}$ of $\mathcal{X}$ can be computed and decomposed as 

\begin{align}
    \bC_{\mathcal{X}} = \frac{1}{n_\textrm{state}-1} \mathcal{X} \mathcal{X}^T = {\bL}_{\mathcal{X}} {\bD}_{\mathcal{X}} {{\bL}_{\mathcal{X}}}^T \label{eq:C}
\end{align}
where the columns of ${\bL}_{\mathcal{X}}$ represent the principal components of $\mathcal{X}$ and ${\bD}_{\mathcal{X}}$ is a diagonal matrix with the corresponding eigenvalues $\{ \lambda_{\mathcal{X},i}, i=0,...,n_\textrm{state}-1\}$ in a decreasing order,
\begin{align}
  {\bD}_{\mathcal{X}} =
  \begin{bmatrix}
    \lambda_{\mathcal{X},0} & & \\
    & \ddots & \\
    & & \lambda_{\mathcal{X},n_\textrm{state}-1}
  \end{bmatrix}.
\end{align}
To compress the state variables to a reduced space of dimension $q \hspace{2mm} (q\in \mathbb{N}^+ \hspace{2mm} \textrm{and} \hspace{2mm} q \leq n_\textrm{state})$, we compute a projection operator ${\bL}_{\mathcal{X},q}$ by keeping the first $q$ columns of ${\bL}_{\mathcal{X}}$. ${\bL}_{\mathcal{X}}$ can be obtained via performing \ac{SVD}~\cite{baker2005singular} which does not require the estimation of the full covariance matrix $\bC_{\mathcal{X}}$. 
For a flattened state field $\bx_t$, the reduced latent vector $\hat{\bx}_t$ reads
\begin{align}
    \hat{\bx}_t =  {{\bL}_{\mathcal{X},q}}^T \bx_t, \label{eq: reconstruction}
\end{align}
which is an approximation to the full vector $\bx_t$. 

The latent vector $\hat{\bx}_t$ can be decompressed to a full space vector $\bx_t^r$ by 
\begin{align}
    \bx_t^r = {{\bL}_{\mathcal{X},q}} \hat{\bx}_t = {{\bL}_{\mathcal{X},q}} \textrm{Tr}({{\bL}_{\mathcal{X},q}}) \bx_t.
\end{align}
The \ac{POD} compression rate $\rho_{\bx}$ and the energy conservation rate $\gamma_{\bx}$ are defined as:
\begin{align}
    \gamma_{\bx} = \sum_{i=0}^{q-1} \lambda^2_{\mathcal{X},i} \Big/ \sum_{i=0}^{n_\textrm{state}-1} \lambda^2_{\mathcal{X},i}  \quad \textrm{and} \quad \rho_{\bx} = q \big/ n_\textrm{state}. \label{eq:POD rate}
\end{align}
To reduce the \ac{DA} computational cost, the assimilation can be carried out in the space of $\hat{\bx}_t$ instead of $\bx_t$, leading to a new state-observation operator $\hat{\mathcal{H}}_t$, defined as
\begin{align}
    \hat{\mathcal{H}}_t = \mathcal{H}_t \circ {{\bL}_{\mathcal{X},q}} \quad \textrm{with} \quad \by_t = \mathcal{H}_t(\bx_t) = \mathcal{H}_t \circ {{\bL}_{\mathcal{X},q}}(\hat{\bx}_t) = \hat{\mathcal{H}}_t(\hat{\bx}_t).
\end{align}
\sibo{It is worth mentioning that the POD here is conducted on uncentered data. This choice is made to retain the original scale and distribution of the data, ensuring a meaningful low-dimensional representation and enabling the data learning mapping function to uncover the underlying physics.}
As for the \ac{ML} inverse operator, instead of applying \ac{POD} to the \ac{CNN} output, we propose an end-to-end \ac{CNN}, named $\text{CNN}_{\text{ROM}}$, from the tessellated observation to the compressed vector $\bx_t$, that is,
\begin{align}
    \tilde{\bY}_t \xrightarrow{\hspace*{0.6cm} \text{CNN}_{\text{ROM}} \hspace*{0.6cm}} \hat{\bx}_t. \label{eq:CNNrom}
\end{align}
The $\text{CNN}_{\text{ROM}}$ follows the classical structure of convolutional encoder with Maxpooling layers~\cite{chen2017deep} as shown in Table~\ref{table: CNN2}. Finally, the objective function of deep data assimilation with \ac{ROM} can be written as
\begin{align}
    J_t(\hat{\textbf{x}})&=\frac{1}{2}\vert \vert\hat{\textbf{x}}-\hat{\textbf{x}}_{b,t}\vert \vert^2_{\hat{\textbf{B}}_t^{-1}}+\frac{1}{2}\vert \vert\hat{\textbf{x}}-\hat{\bx}_{v,t}\vert \vert^2_{\hat{\textbf{P}}_t^{-1}}+\frac{1}{2}\vert \vert\textbf{y}_t-\hat{\mathcal{H}}_t(\hat{\textbf{x}})\vert \vert^2_{\textbf{R}_t^{-1}}, \label{eq:DA_ROM}
\end{align}
where 
\begin{align}
   \hat{\bx}_{v,t} = \text{CNN}_{\text{ROM}} (\tilde{\bY}_t) \quad \textrm{and} \quad \hat{\textbf{B}}_t = {{\bL}_{\mathcal{X},q}}^T {\textbf{B}}_t {{\bL}_{\mathcal{X},q}}.
\end{align}
The minimization of Equation~\eqref{eq:DA_ROM} can be performed through Algorithm~\ref{algo:2} with $(\hat{\bx}_{b,t}, \by_t, \hat{\bB}_t, \hat{\bR}_t, \hat{\bP}_t,\hat{\mathcal{H}}_t)$ as inputs instead of $(\bx_{b,t}, \by_t, \bB_t, \bR_t, \bP_t,\mathcal{H}_t)$ .
\begin{table}[h!]
\centering
\caption{NN structure of the \ac{CNN} with $\tilde{\bY}_t$ as input and $\hat{\bx}_t$ as output}
\begin{tabular}{ccc} \toprule
    {\textbf{Layer (type)}} 
    & {\textbf{Output Shape}} 
    & {\textbf{Activation}} \\ \midrule
     & &  \\
    {Input}  
    & {$(N_y, M_y,1)$}
    & {}\\
    {\textrm{Conv 2D} $(8\times8)$}  
    & {$(N_y, M_y,16)$}
    & {ReLu}\\
    {\textrm{Conv 2D} $(8\times8)$}  
    & {$(N_y, M_y,16)$}
    & {ReLu}\\
    {\textrm{Maxpooling} $(2\times2)$}  
    & {$(\lceil N_y/2 \rceil, \lceil M_y/2 \rceil,16)$}
    & {}\\
    {\textrm{Conv 2D} $(4\times4)$}  
    & {$(\lceil N_y/2 \rceil, \lceil M_y/2 \rceil,16)$}
    & {ReLu}\\
    {\textrm{Maxpooling} $(2\times2)$}  
    & {$(\lceil N_y/4 \rceil, \lceil M_y/4 \rceil,16)$}
    & {}\\
    {\textrm{Conv 2D} $(4\times4)$}  
    & {$(\lceil N_y/4 \rceil, \lceil M_y/4 \rceil,16)$}
    & {ReLu}\\
    {\textrm{Flatten}}  
    & {$(\lceil N_y/4 \rceil \times \lceil M_y/4 \rceil \times 16)$}
    & {}\\
    {\textrm{Dense}(q)}  
    & {$q$}
    & {}\\ 
    \bottomrule
\end{tabular}
\label{table: CNN2}
\end{table}

\section{Numerical experiments}

\label{sec:experiments}

In this section, we illustrate the numerical results in a two-dimensional \ac{CFD} system. The proposed deep \ac{DA} approach is compared against conventional \ac{DA} and the \ac{VCNN} approach. We show the robustness of \ac{VIVID} against varying prior error levels, different sensor numbers and error covariance misspecifications. Two metrics, namely \ac{R-RMSE} and \ac{SSIM} are used to evaluate the difference and the similarity between the assimilated and the true state, respectively.

\subsection{Description of the shallow water system and experiments set up}
Here we consider a standard shallow-water fluid mechanics system which is broadly used for evaluating the performance of \ac{DA} approaches (e.g.,~\cite{Heemink1995},~\cite{Cioaca2014},~\cite{cheng2021error}). We consider a non-linear wave-propagation problem. The initial condition consists of a cylinder of water with a certain radius that is released at  $\textit{t}=0$. Here, we assume that the horizontal length scale is more important than the vertical one. The Coriolis force is also neglected. These lead to the following Saint-Venant equations (\cite{Venant1871}) coupling the horizontal fluid velocity and height:

\begin{SCfigure}[][h]
  \begin{minipage}{.5\textwidth}
\begin{align}
    \frac{\partial u}{\partial t}&=-g\frac{\partial}{\partial x}(h)-bu \label{eq: sw} \\
    \frac{\partial v}{\partial t}&=-g\frac{\partial}{\partial y}(h)-bv  \notag\\ 
    \frac{\partial h}{\partial t}&=-\frac{\partial}{ \partial x}(uh)-\frac{\partial}{\partial y}(v h)  \notag \\
    u_{t=0} &= 0 \notag \\
    v_{t=0} &= 0 \notag 
\end{align}
  \end{minipage}
    \begin{minipage}[H]{.0\textwidth}
      \includegraphics[width=2.in]{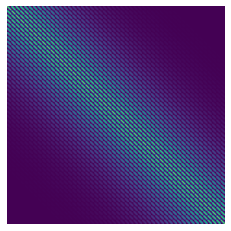}
      \\
      \\
      \caption{The error covariance matrix with scale correlation length $L=5$ for a flattened state vector}
      \label{fig:balgo}
  \end{minipage}
\end{SCfigure}
 where $(u,v)$ denote the two dimensional fluid velocity (in $0.1m/s$) and $h$ represents the fluid height (in $mm$). The earth gravity constant $g$ is thus scaled to 1 and the dynamical system is set in a non-conservative form. The initial velocity fields are set to zero for both $u$ and $v$. The height of the water cylinder is set as $h_p$ $mm$ higher than the still water and the radius of the initial water cylinder is set to be $r_w$ $mm$. $h_p$ and $r_w$ are considered as hyperparameters in this modelling. The study area of $(50mm \times 50mm)$ is discretized with a squared grid of size $(N_x = 50, M_x = 50)$ and the solution of Eq. (\ref{eq: sw}) is approximated using a first order finite difference method. The time integration is also performed using a finite difference scheme with a time interval $\delta t=10^{-6} s$. 

\begin{figure*}[ht!]
\centering
\makebox[\linewidth][c]{
\includegraphics[width = 1in]{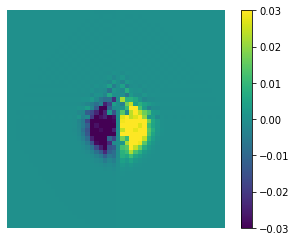}
\includegraphics[width = 1in]{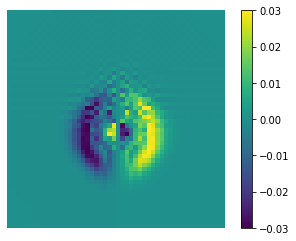}
\includegraphics[width = 1in]{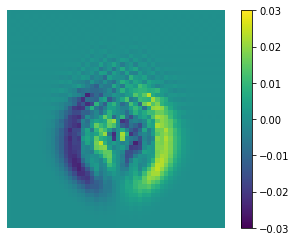}
\includegraphics[width = 1in]{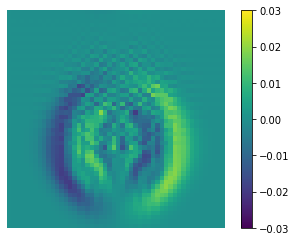}
\includegraphics[width = 1in]{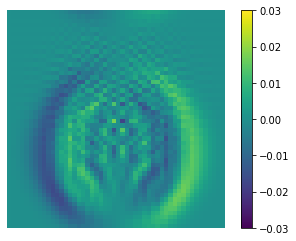}
\includegraphics[width = 1in]{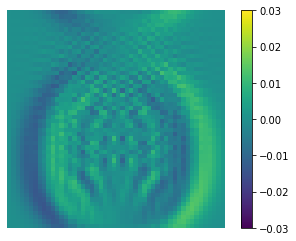}}\\
\makebox[\linewidth][c]{
\includegraphics[width = 1in]{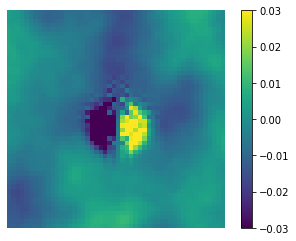}
\includegraphics[width = 1in]{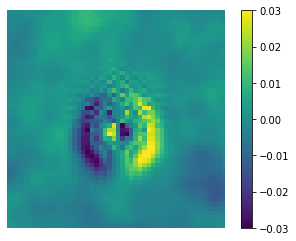}
\includegraphics[width = 1in]{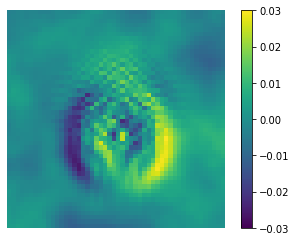}
\includegraphics[width = 1in]{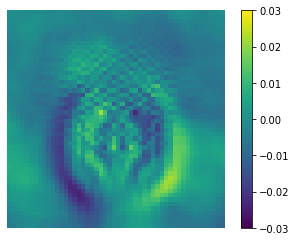}
\includegraphics[width = 1in]{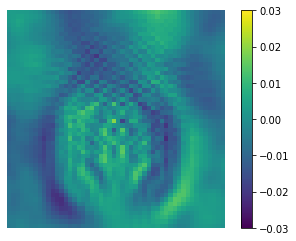}
\includegraphics[width = 1in]{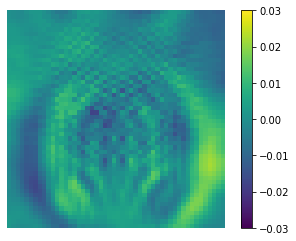}}\\
\makebox[\linewidth][c]{
\includegraphics[width = 1in]{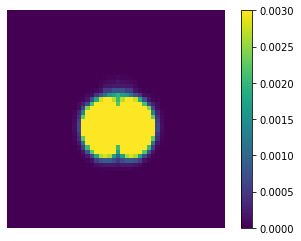}
\includegraphics[width = 1in]{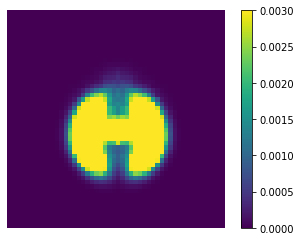}
\includegraphics[width = 1in]{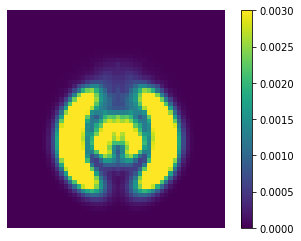}
\includegraphics[width = 1in]{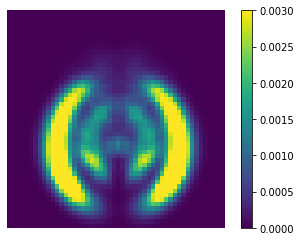}
\includegraphics[width = 1in]{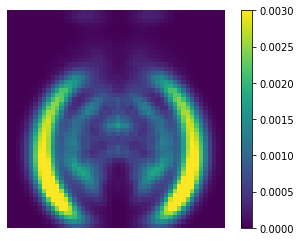}
\includegraphics[width = 1in]{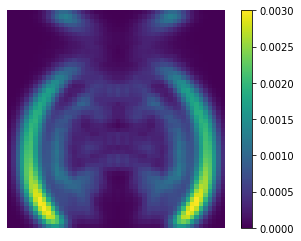}}\\
\makebox[\linewidth][c]{
\includegraphics[width = 1in]{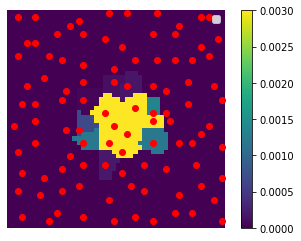}
\includegraphics[width = 1in]{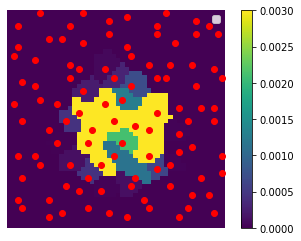}
\includegraphics[width = 1in]{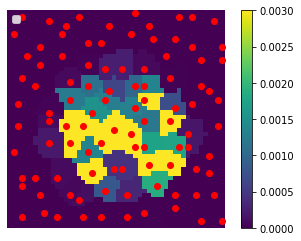}
\includegraphics[width = 1in]{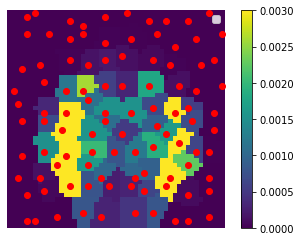}
\includegraphics[width = 1in]{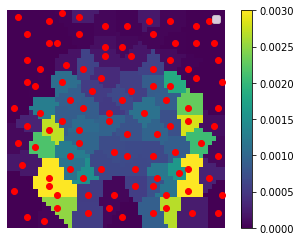}
\includegraphics[width = 1in]{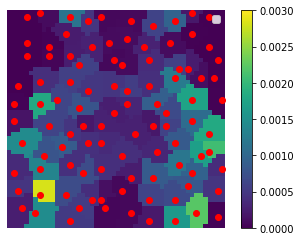}}\\
\caption{The evolution of the simulated velocity field (1st raw), the background velocity field (2nd raw), the complete observation space (3rd raw) and the tessllated observation field with 100 sensors (4th raw) at $\{5, 10, 15, 20, 25, 30 \}\times 10^{-4}s $ respectively.}
\label{fig:fields}
\end{figure*}

 In this study, we focus on the reconstruction of the velocity field $u$ based on sparse and time-varying sensors and noisy prior forecasts. For the latter, Gaussian spatial-correlated noises are added to the simulated $u$ field, more precisely, a Matérn covariance kernel function of order $3/2$ is employed, 
\begin{align}
    \phi(r)=(1+\frac{r}{L}) \exp(-\frac{r}{L}), \label{eq:balgo}
\end{align}
where $r,L$ are the spatial distance and the correlation scale length defined in Equation~\eqref{eq:Gaspari}. In other words, 
\begin{align}
    \bx_t = u_t, \quad \bx_{b,t} = \bx_t + s_b \cdot \epsilon_{b,t} \quad \textrm{and} \quad \epsilon_{b,t} \sim \mathcal{N}(0,\bB(\phi(L,r)),
\end{align}
where $s_b$ is the background error standard deviation and $\bB(\phi(r))$ represents the error correlation matrix of the flattened velocity field $\bx_{b,t}$. \sibo{The subscript ${(.)}_b$ refers to the background (prior) information.}
In this paper, $\bB(\phi(L,r))$ is fixed with $L=5$ as shown in Figure~\ref{fig:balgo}. To evaluate the performance of the proposed assimilation approach in complex non-linear systems, a synthetic observation function is created to build the observation space $\bY_t$. Different sensors  $\{y_{t,k}\}_{k=1...k^*} $ with a random placement are chosen at each time step $t$. The details of the non-linear transformation function and the sensor placement choice are available in the appendix of this paper. An example of the dynamical velocity field with the associated background field, the full observation field and the tessllated observations (with 100 sensors) are illustrated in Figure~\ref{fig:fields}. In this example, we fix $h_p= 0.1mm, r_w= 4mm$ in the simulation and the background error standard deviation is set to be $0.005$ (in $0.1m/s$). 

\begin{figure*}[h!]
\centering
\includegraphics[width=\textwidth]{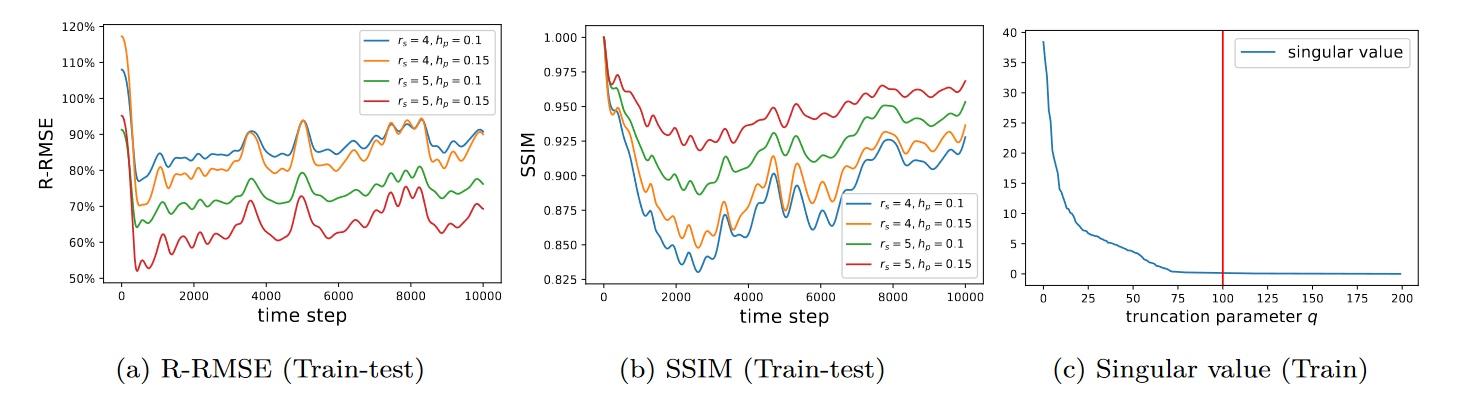}
\caption{R-RMSE (a) and SSIM (b) between the train (4 different simulations) and the test data (1 simulation); Singular value of all the train data (c)}
\label{fig:testdiff}
\end{figure*}

As for the \ac{VCNN} inverse operator, the train data consists of four simulations and the tests of both \ac{VCNN} and \ac{VIVID} are performed with an unseen simulation, as shown in Table~\ref{table: traintest}. Each simulation consists of 10000 snapshots from $0s$ to $10^{-2}s$. The test dataset is significantly different from the train dataset, resulting in an average \ac{R-RMSE} of 82.5\% and an average \ac{SSIM} of 0.917 as shown in Figure~\ref{fig:testdiff} (a,b).

The \ac{VCNN} operator is trained on the Google Colab server with an NVIDIA T4 GPU. For a fair comparison, all \ac{DA} approaches are \sibo{performed on the same labtop CPU of Intel(R) Core(TM) i7-10810U with 16 GB Memory.} The minimization of the objective function in \ac{DA} (Equation~\eqref{eq_3dvar} and~\eqref{eq:DA_op}) is carried out using the BFGS approach with a stopping criteria~\cite{blelly2018stopping} of tolerance $10^{-6}$. The truncation parameter $q$ (i.e., the dimension of the reduced space) is chosen as $q=100$ where the singular values reach a point of stagnation as shown in Figure~\ref{fig:testdiff}~(c).

\begin{table}[h!]
\centering
\caption{Model parameter choice in the train and test datasets}
\begin{tabular}{c|cccc|c} \toprule
    {Parameter} 
    & \multicolumn{4}{c|}{Train} 
    & test \\ \midrule
    $h_p$  
    & 0.1 & 0.15 & 0.1 & 
     0.15 & 0.2\\
    {$r_w$}  
    & 4 & 4 & 5 & 5 
    & 6\\
    \bottomrule
\end{tabular}
\label{table: traintest}
\end{table}

\subsection{Numerical results and analysis of the shallow water experiments}
In this section, we show extensive experiments to evaluate the performance of the proposed approach against conventional \ac{DA} and only \ac{VCNN} approaches, regarding both prediction accuracy and efficiency. In the test dataset, \ac{DA} is performed on 20 snapshots (extracted every $5 \times 10^{-4} s$) at different time steps. The simulation outputs are considered as ground truth, where spatially correlated errors are added to construct background states. In each \ac{DA}, the sensor placement is randomly chosen with a given sensor number (see appendix). 
\subsubsection{Accuracy and efficiency} 
We first set the background error standard deviation to $0.02m/s$, leading to an \ac{R-RMSE} of $0.62$ and a \ac{SSIM} of 0.901 in average for all test samples. As shown in Figure~\ref{fig:fields}, 100 time-varying sensors are placed randomly (with a given structure) in both train and test snapshots. The observations are first set to be error-free. Thus the observation error matrix is set to be $\bR_t = 10^{-3}\cdot \bI$ to increase the confidence level of the observed data, where $\bI$ denotes the identity matrix. The inverse operator covariance $\bP_t$ is empirically estimated and localized, as explained in Equation~\eqref{eq:ensemble_P}-\eqref{eq:Gasp}. The $\bB_t$ matrix with the Matern covariance kernel (Eqaution~\eqref{eq:balgo}) is supposed to be perfectly known here. 

\begin{figure}[h!]
\centering
\subfloat[t=$0.04s$]{\includegraphics[width = 1.7in]{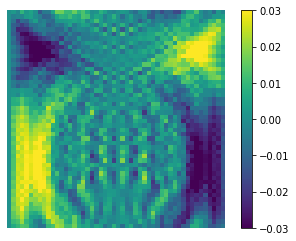}}
\subfloat[t=$0.08s$]{\includegraphics[width = 1.7in]{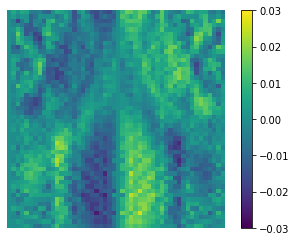}}
\caption{Simulated velocity field, considered as the ground truth in the numerical experiments}
\label{fig:truth}
\end{figure}

\begin{figure}[h!]
  \centering
\includegraphics[width=\textwidth]{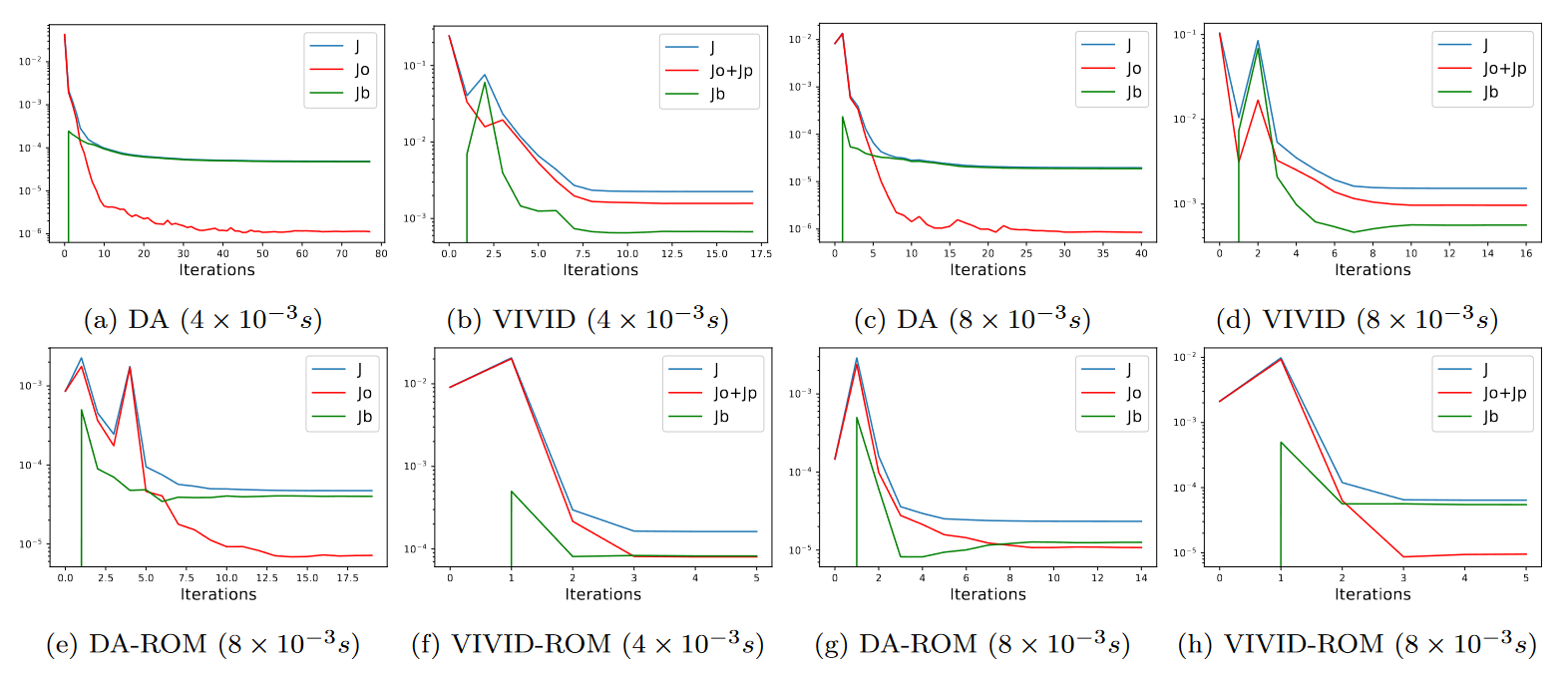}
\caption{Minimization (L-BFGS) iterations in variational DA }
\label{fig:L-BFGS_iter}
\end{figure}

\begin{figure}[h!]
\centering
\subfloat[background ]{\includegraphics[width = 1.7in]{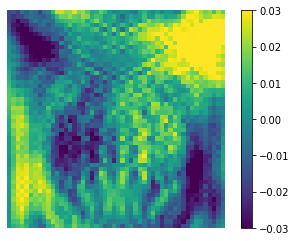}}
\subfloat[DA]{\includegraphics[width = 1.7in]{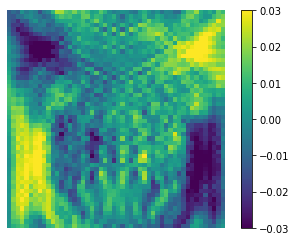}}
\subfloat[VCNN]{\includegraphics[width = 1.7in]{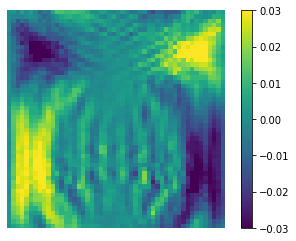}}
\subfloat[VIVID ]{\includegraphics[width = 1.7in]{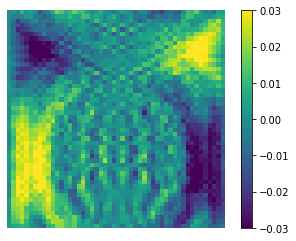}}
\\
\subfloat[error background]{\includegraphics[width = 1.7in]{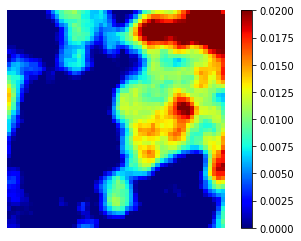}}
\subfloat[error DA]{\includegraphics[width = 1.7in]{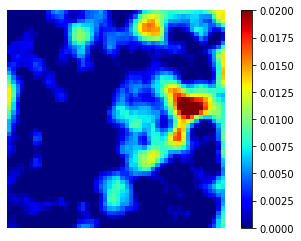}}
\subfloat[error VCNN]{\includegraphics[width = 1.7in]{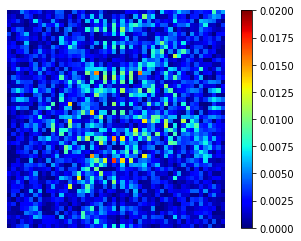}}
\subfloat[error VIVID ]{\includegraphics[width = 1.7in]{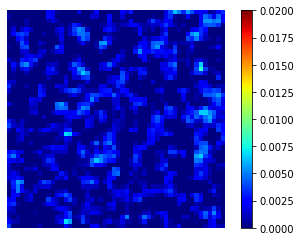}}\\

\subfloat[background ]{\includegraphics[width = 1.7in]{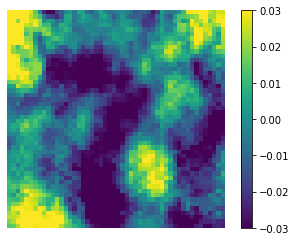}}
\subfloat[DA]{\includegraphics[width = 1.7in]{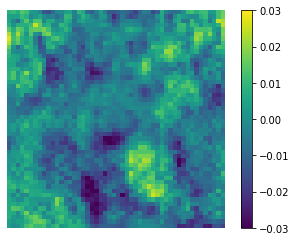}}
\subfloat[VCNN]{\includegraphics[width = 1.7in]{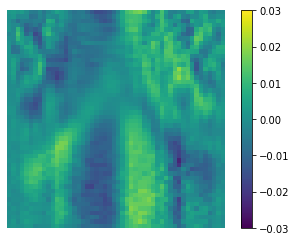}}
\subfloat[VIVID ]{\includegraphics[width = 1.7in]{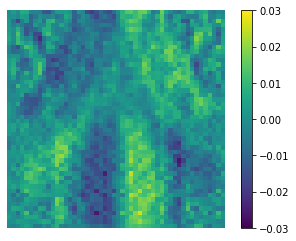}}\\
\subfloat[error background]{\includegraphics[width = 1.7in]{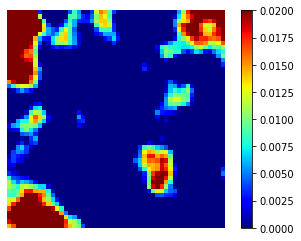}}
\subfloat[error DA]{\includegraphics[width = 1.7in]{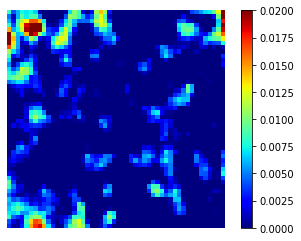}}
\subfloat[error VCNN]{\includegraphics[width = 1.7in]{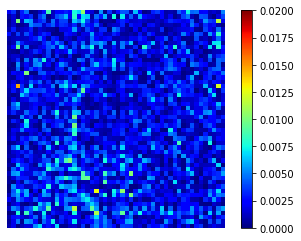}}
\subfloat[error VIVID ]{\includegraphics[width = 1.7in]{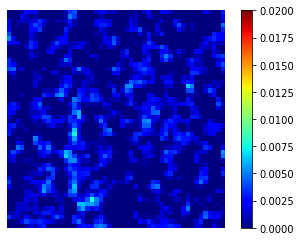}}\\
   \caption{Background and assimilated velocity field at $t=0.04s$ (a-h) and $t=0.08s$ (i-p) compared to the ground truth}
   \label{fig:res1}
\end{figure}

\begin{figure}[h!]
\subfloat[background ]{\includegraphics[width = 1.7in]{Figures/Xb_4000_final.png}}
\subfloat[DA (ROM)]{\includegraphics[width = 1.7in]{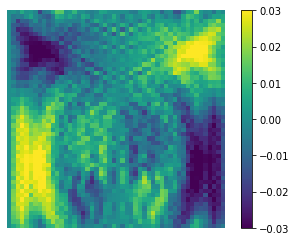}}
\subfloat[VCNN (ROM)]{\includegraphics[width = 1.7in]{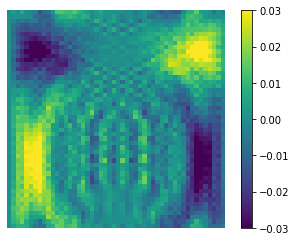}}
\subfloat[VIVID (ROM)]{\includegraphics[width = 1.7in]{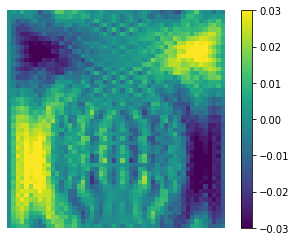}}\\
\subfloat[error background]{\includegraphics[width = 1.7in]{Figures/Xb_4000_error_final.png}}
\subfloat[error DA]{\includegraphics[width = 1.7in]{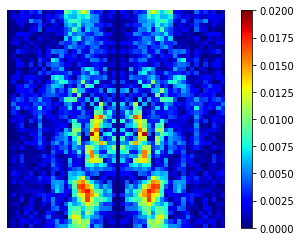}}
\subfloat[error VCNN]{\includegraphics[width = 1.7in]{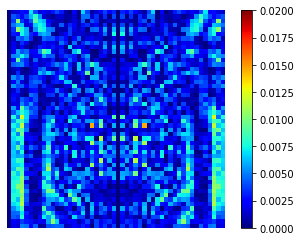}}
\subfloat[error VIVID ]{\includegraphics[width = 1.7in]{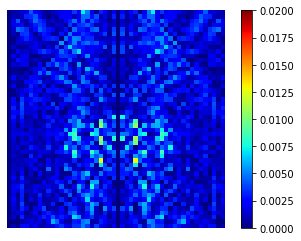}}\\

\subfloat[background ]{\includegraphics[width = 1.7in]{Figures/Xb_8000_final.png}}
\subfloat[DA (ROM)]{\includegraphics[width = 1.7in]{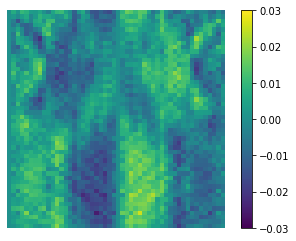}}
\subfloat[VCNN (ROM)]{\includegraphics[width = 1.7in]{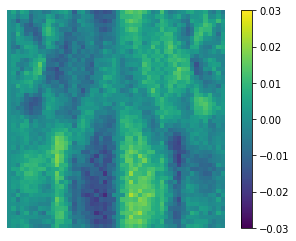}}
\subfloat[VIVID (ROM)]{\includegraphics[width = 1.7in]{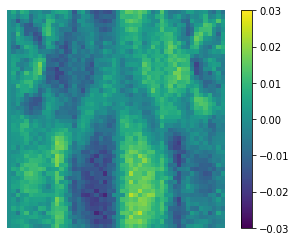}}\\
\subfloat[error background]{\includegraphics[width = 1.7in]{Figures/Xb_8000_error_final.png}}
\subfloat[error DA]{\includegraphics[width = 1.7in]{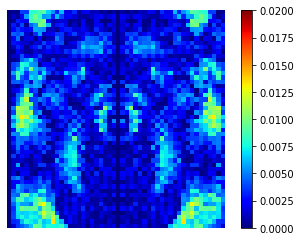}}
\subfloat[error VCNN]{\includegraphics[width = 1.7in]{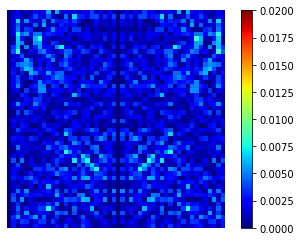}}
\subfloat[error VIVID ]{\includegraphics[width = 1.7in]{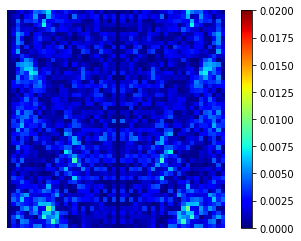}}\\
   \caption{Background and assimilated (with ROM) velocity field at $t=0.04s$ (a-h) and $t=0.08s$ (i-p) compared to the ground truth}
   \label{fig:res2}
\end{figure}

As illustrated in Table~\ref{tab:results}, all \ac{DA} and \ac{DL} methods can significantly reduce the \ac{R-RMSE} and increase the \ac{SSIM}. In particular, the proposed \ac{VIVID} approach substantially outperforms conventional \ac{DA} and \ac{VCNN} in terms of assimilation accuracy. Furthermore, we find that \ac{VIVID} can considerably reduce the computational time by decreasing the number of iterations required in the L-BFGS optimization for variational assimilation (see Algorithm~\ref{algo:1} and~\ref{algo:2}). As observed in~\cite{frerix2021variational}, the improvement of efficiency is due to the extra information provided by the \ac{DL} inverse operator which links the observations directly to the state variables. In other words, the minimization loops can be better guided towards the observations in \ac{VIVID}. The evolution of the \ac{DA} objective functions against L-BFGS iterations is illustrated in Figure~\ref{fig:L-BFGS_iter}. It can be clearly noticed that the objective function of \ac{VIVID} stabilizes much faster compared to the conventional \ac{DA} both in the full and reduced space. It is worth mentioning that, unlike \ac{DL} loss functions, the value of the objective function in \ac{DA} does not necessarily represent the assimilation accuracy since the ground truth is not involved in \ac{DA} objective functions. 
\begin{table}[h!]
\centering
\caption{}
\begin{tabular}{c|cc|cc} \toprule 
    {Method} 
    & \multicolumn{2}{c|}{Accuracy} 
    & \multicolumn{2}{c}{Efficiency}  \\ \midrule
      \midrule
    & R-RMSE & SSIM & iterations & time \\
    background  
    & 0.62 & 0.901 & - & -  \\ \midrule
    DA  
    & 0.33 & 0.979 & 54.5 & $\approx 42\text{min}$ \\
    VCNN  
    &  0.32 & 0.993 & - & - \\
    VIVID 
    & 0.14  & 0.998 &  15.1 & $\approx 11\text{min}$\\ \midrule
    DA-ROM  
    &  0.49 & 0.984 & 19.1 & $\approx 8s$  \\
    VCNN-ROM  
    & 0.33 & 0.992 & - & -  \\
    VIVID-ROM  
    & 0.16 & 0.993 & 5.2 & $\approx 2s$ \\
    \bottomrule
\end{tabular}
\label{tab:results}
\end{table}
Figure~\ref{fig:res1} illustrates the reconstructed velocity field at $t=0.04s$ and $t=0.08s$ with the mismatch against the ground truth (Figure~\ref{fig:truth}).  As shown by Figure~\ref{fig:res1} (f,n), the capability of \ac{DA} is limited where the prior error is substantial and spatially correlated. On the other hand, \ac{VCNN} manages to provide a good overall reconstruction (Figure~\ref{fig:res1} (c,k)), leading to a high \ac{SSIM} in Table~\ref{tab:results}. However, the reconstruction noise is still significant (see Figure~\ref{fig:res1} (g,o)) due to the time-varying sensors and the difference between train and test datasets (see Table~\ref{table: traintest}). In contrast, \ac{VIVID}, by definition, combining the strength of conventional \ac{DA} and \ac{VCNN}, provides an accurate overall field reconstruction, which is consistent with the observation in Table~\ref{tab:results}. The same experiments, coupled with \ac{ROM}, \sibo{are shown} in Figure~\ref{fig:res2}. We find that the application of \ac{POD} can not only reduce the computational cost (see Table~\ref{tab:results}) but also denoise the added background error. This is mainly because the constructed principle components have learned underlying physics information~\cite{jha2010denoising} despite the difference between train and test datasets. It is worth noting that the \ac{ROM} has well understood the left-right symmetricity in the velocity field, resulting in symmetric reconstructed fields in \ac{DA}, \ac{VCNN} and \ac{VIVID}. In comparison to the other two approaches, \ac{VIVID} delivers the most accurate velocity field (Figure~\ref{fig:res2}), leading to the best \ac{R-RMSE} and \ac{SSIM} score in Table~\ref{tab:results}. 

In the previous experiments, \ac{VIVID} shows outstanding performance both when being applied directly in the full physical space or coupled with a \ac{ROM}.
From now on, to facilitate direct comparison, we only apply different approaches in the full velocity field at $t=0.04s$ and evaluate their robustness against varying prior errors, sensor numbers and misspecified background error matrices.

\subsubsection{Varying background error}
\begin{figure}[h!]
  \centering
\includegraphics[width=0.75\textwidth]{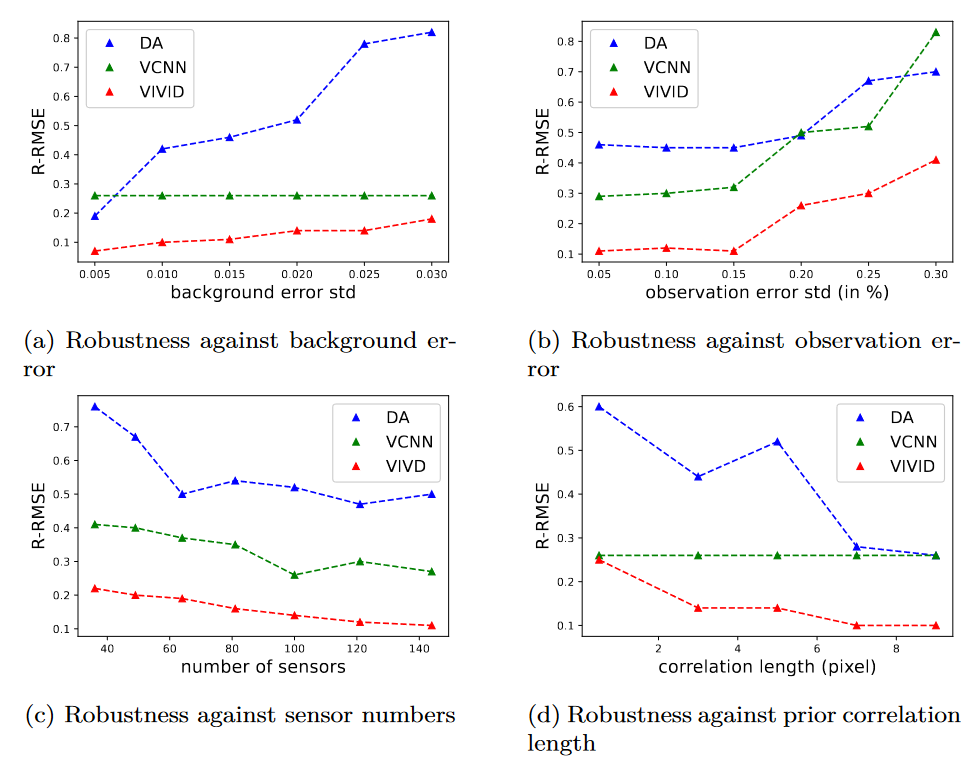}
\caption{Evolution of R-RMSE according to different assimilation assumptions}
\label{fig:evolution}
\end{figure}

\begin{figure}[h!]
\subfloat[background ]{\includegraphics[width = 1.7in]{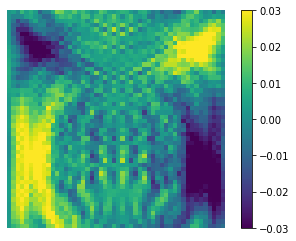}}
\subfloat[DA]{\includegraphics[width = 1.7in]{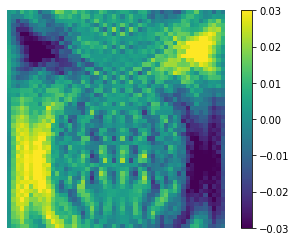}}
\subfloat[VCNN]{\includegraphics[width = 1.7in]{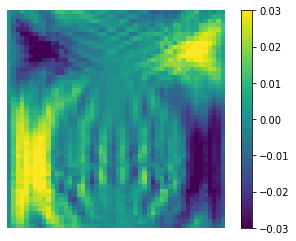}}
\subfloat[VIVID ]{\includegraphics[width = 1.7in]{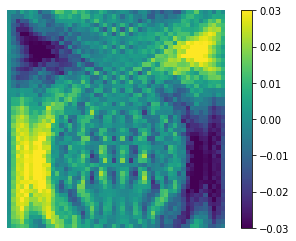}}\\
\subfloat[error background]{\includegraphics[width = 1.7in]{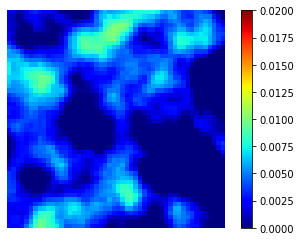}}
\subfloat[error DA]{\includegraphics[width = 1.7in]{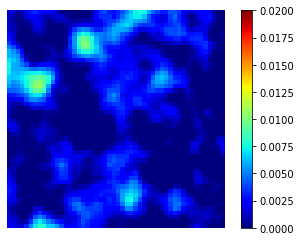}}
\subfloat[error VCNN]{\includegraphics[width = 1.7in]{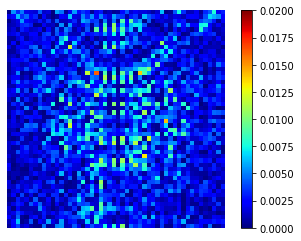}}
\subfloat[error VIVID ]{\includegraphics[width = 1.7in]{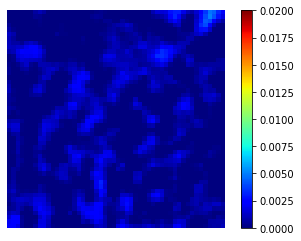}}\\
   \caption{Background and assimilated velocity field at $t=0.04s$ with a background error of standard deviation of 
   $0.005m/s$}
   \label{fig:res3}
\end{figure}

With the same spatial error correlation defined in Equation~\eqref{eq:balgo}, we first vary the background error deviation from $(0.005,0.01,0.015,0.02,0.025,0.03)$ (in $m/s$). We plot the evolution of the reconstruction error in Figure~\ref{fig:evolution} (a). As expected, the assimilation error of \ac{DA} and \ac{VIVID} increase against the background error standard deviation. However, \ac{VIVID} is much less sensitive to the background error thanks to the \ac{VCNN} inverse operator, which is not impacted by $\bx_b$. The experiments with a small background error deviation ($0.005m/s$) are illustrated in Figure~\ref{fig:res3}. As observed, low background error (Figure~\ref{fig:res3} (f)) leads to accurate assimilation results (Figure~\ref{fig:res3} (g,h)). Despite that conventional \ac{DA} outperforms \ac{VCNN} when the background error is small, the proposed \ac{VIVID}, combining the advantage of \ac{DA} and \ac{DL}, shows a significant advantage in reconstruction accuracy compared to the other two approaches.   

\subsubsection{Varying observation error}
\begin{figure}[h!]
\subfloat[background ]{\includegraphics[width = 1.7in]{Figures/Xb_4000_final.png}}
\subfloat[DA]{\includegraphics[width = 1.7in]{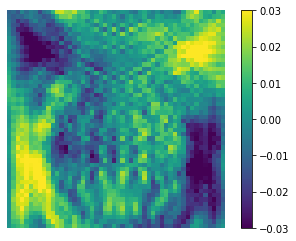}}
\subfloat[VCNN]{\includegraphics[width = 1.7in]{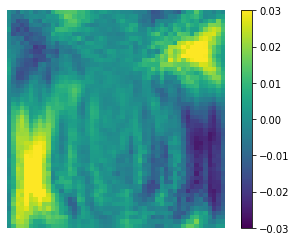}}
\subfloat[VIVID ]{\includegraphics[width = 1.7in]{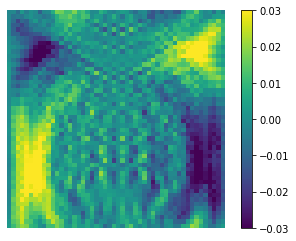}}
\
\subfloat[error background]{\includegraphics[width = 1.7in]{Figures/Xb_4000_error_final.png}}
\subfloat[error DA]{\includegraphics[width = 1.7in]{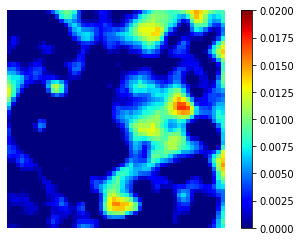}}
\subfloat[error VCNN]{\includegraphics[width = 1.7in]{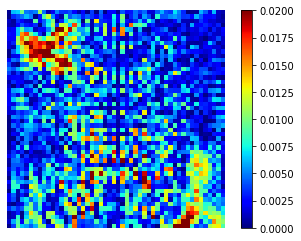}}
\subfloat[error VIVID ]{\includegraphics[width = 1.7in]{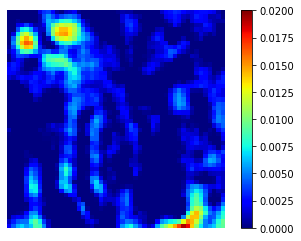}}\\
   \caption{Background and assimilated velocity field at $t=0.04s$ with $30\%$ of observation error}
   \label{fig:res4}
\end{figure}

Until now, the observations are assumed error-free. To further explore the robustness of different approaches, spatial-independent observation errors are now added to the observation vector $\by$. These observation errors are supposed to be relative, varying from $5\%$ to $30\%$, regarding the true observation value. As observed in Figure~\ref{fig:evolution} (b), all methods are sensitive to the observation error, in particular \ac{VCNN} which is trained with error-free observation data. Nonetheless, \ac{VIVID} shows a consistent advantage in terms of R-RMSE regardless of the observation error. The assimilated fields with a relatively high observation error ($30\%$) is shown in Figure~\ref{fig:res4}. Compared to error-free observations, suboptimal reconstruction results are observed in all three methods compared to Figure~\ref{fig:res1}. Despite achieving the most accurate field reconstruction, \ac{VIVID} is clearly impacted by the noise in both conventional \ac{DA} and \ac{VCNN}. 

\subsubsection{Varying sensor numbers}

\begin{figure}[h!]
\subfloat[background ]{\includegraphics[width = 1.7in]{Figures/Xb_4000_final.png}}
\subfloat[DA]{\includegraphics[width = 1.7in]{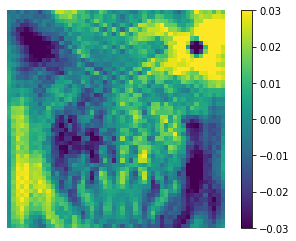}}
\subfloat[VCNN]{\includegraphics[width = 1.7in]{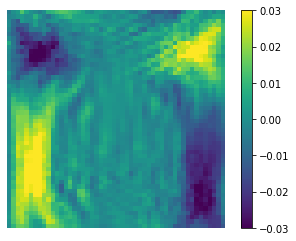}}
\subfloat[VIVID ]{\includegraphics[width = 1.7in]{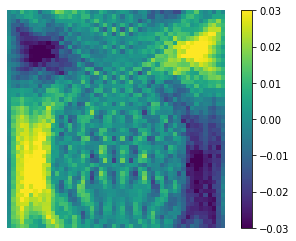}}\\
\subfloat[error background]{\includegraphics[width = 1.7in]{Figures/Xb_4000_error_final.png}}
\subfloat[error DA]{\includegraphics[width = 1.7in]{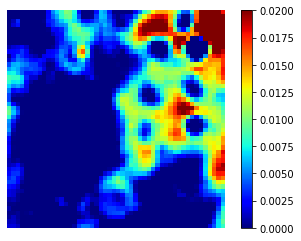}}
\subfloat[error VCNN]{\includegraphics[width = 1.7in]{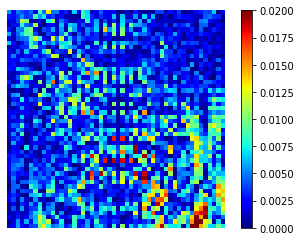}}
\subfloat[error VIVID ]{\includegraphics[width = 1.7in]{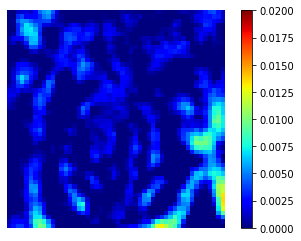}}\\

\subfloat[background ]{\includegraphics[width = 1.7in]{Figures/Xb_4000_final.png}}
\subfloat[DA]{\includegraphics[width = 1.7in]{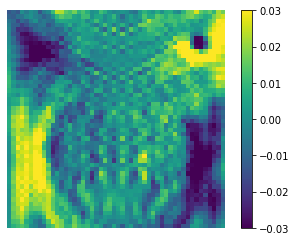}}
\subfloat[VCNN]{\includegraphics[width = 1.7in]{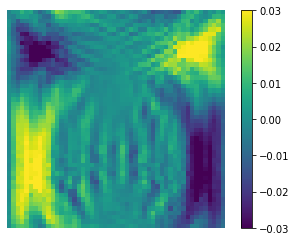}}
\subfloat[VIVID ]{\includegraphics[width = 1.7in]{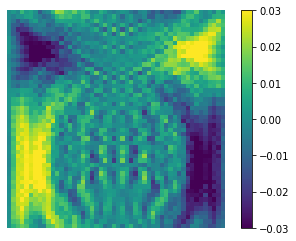}}\\
\subfloat[error background]{\includegraphics[width = 1.7in]{Figures/Xb_4000_error_final.png}}
\subfloat[error DA]{\includegraphics[width = 1.7in]{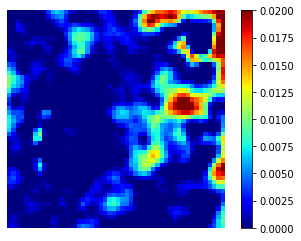}}
\subfloat[error VCNN]{\includegraphics[width = 1.7in]{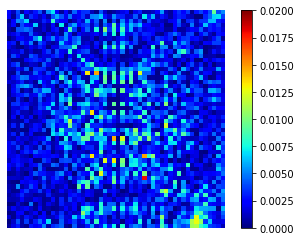}}
\subfloat[error VIVID ]{\includegraphics[width = 1.7in]{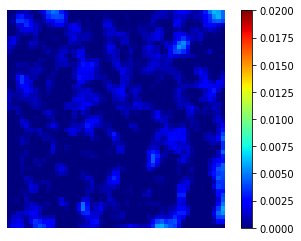}}\\
   \caption{Background and assimilated velocity field at $t=0.04s$ with 36 (a-i) and 144 (a-i) sensors}
   \label{fig:res5}
\end{figure}

We implement the experiments with various numbers of sensors. The sensors are randomly placed within a certain range $r_s = 3$ (see appendix) to the standard placement (i.e., $6 \times 6, 7 \times 7, ...,12 \times 12$ in this section). Clearly, a large number of sensors provide more observation information to the assimilation. On the other hand, \ac{VCNN} is solely trained with $10 \times 10 = 100$ sensors. To ensure the generalizability of the proposed approach, it is thus important to assess the robustness with different numbers of sensors. Figure~\ref{fig:evolution} (c) illustrates the evolution of R-RMSE. It can be clearly observed that overall the R-RMSE decreases against the number of sensors, in particular, the R-RMSE of \ac{VIVID} drops $50\%$ from 0.2 to 0.1. We illustrate the experiments with $36$ and $144$ sensors for comparison in Figure~\ref{fig:res5}. Compared to Figure~\ref{fig:res1} (with 100 sensors), as expected, using only 36 sensors leads to suboptimal assimilation results due to the sparsity of observations. Furthermore, comparing Figure~\ref{fig:res5} (o) and Figure~\ref{fig:res1} (g), we can conclude that using 144 sensors also perturbs \ac{VCNN} trained on data with 100 sensors. However, the different number of sensors only slightly impact the performance of \ac{VIVID} as shown in Figure~\ref{fig:res5} (h,p).    

\subsubsection{misspecified covariance matrix}
Misspecified error covariances can lead to suboptimal assimilation results~\cite{tandeo2020review}. Estimating these covariance matrices, especially the background matrix, has been a long-standing research challenge~\cite{tandeo2020review, valler2019impact,cheng2022observation}. In this paper, we study the impact of misspecifying the background error correlation length $L$ defined in Equation~\eqref{eq:balgo}. The background states in the numerical experiments are all generated with $L=5$
in this paper. Here, we vary the estimated correlation length $L^E$ with $L^E \in \{ 0.1,3,5,7,9\}$ (in pixel length). Notably, $L^E=0.1$ signifies that almost no spatial error correlation is considered in the assimilation procedure. 

Figure~\ref{fig:evolution} (d) depicts the influence of $L^E$ on assimilation R-RMSE knowing the exact correlation length is $L=5$. Underestimating the correlation length ($L^E << L$) clearly leads to a high R-RMSE, indicating suboptimal assimilation for both conventional \ac{DA} and \ac{VIVID}. Surprisingly we find that overestimating the correlation length results in a more accurate field reconstruction for both \ac{DA} and \ac{VIVID}. The R-RMSE of \ac{VIVID} stabilizes around 0.1 for $L^E > 3$.

\begin{figure}[h!]
\subfloat[DA (L=0.1)]{\includegraphics[width = 1.7in]{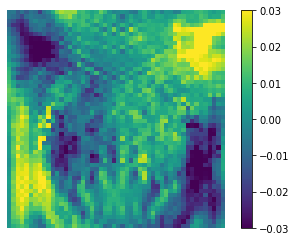}}
\subfloat[VIVID (L=0.1)]{\includegraphics[width = 1.7in]{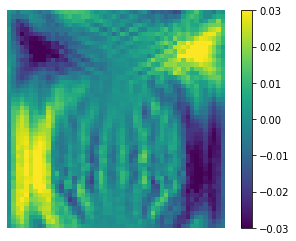}}
\subfloat[DA (L=9)]{\includegraphics[width = 1.7in]{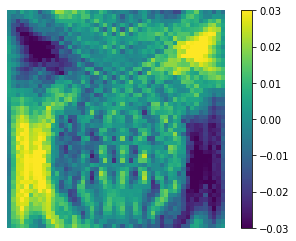}}
\subfloat[VIVID (L=9)]{\includegraphics[width = 1.7in]{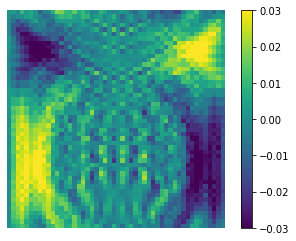}}
\\
\subfloat[error DA]{\includegraphics[width = 1.7in]{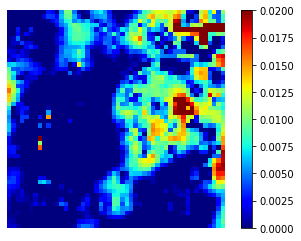}}
\subfloat[error VIVID]{\includegraphics[width = 1.7in]{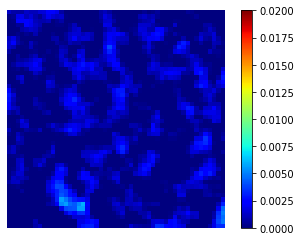}}
\subfloat[error DA]{\includegraphics[width = 1.7in]{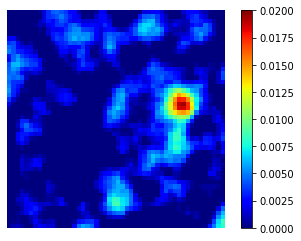}}
\subfloat[error VIVID ]{\includegraphics[width = 1.7in]{Figures/Xa_proposed_4000_error_cor9.png}}\\
   \caption{Background and assimilated velocity field at $t=0.04s$ with a background error of standard deviation of 
   $0.005m/s$. The VCNN results are the same as in Figure~\ref{fig:res1}}
   \label{fig:res6}
\end{figure}

Figure~\ref{fig:res6} displays the assimilation outputs with $L^E=0.1$ and $L^E=9$. As shown in Figure~\ref{fig:res6} (a,e), the underestimation of the correlation length leads to insufficient use of the observation information, especially in conventional \ac{DA}. Consistent with Figure~\ref{fig:evolution}, overestimation of the correlation length, in contrast, results in less assimilation error. In both cases, an advantage of \ac{VIVID} is clearly identified. These results demonstrate the strength of \ac{VIVID} when the background error is misspecified, which is applicable for a wide range of real-world \ac{DA} problems. 

In summary, the results in this section with extensive numerical experiments show the consistent advantage of \ac{VIVID} regarding the different levels of background error, observation error, number of sensors and misspecification of error covariances. 

\section{Discussion and future work}
\label{sec:discuss}

\begin{figure*}[h!]
\centering
\includegraphics[width=0.45\textwidth]{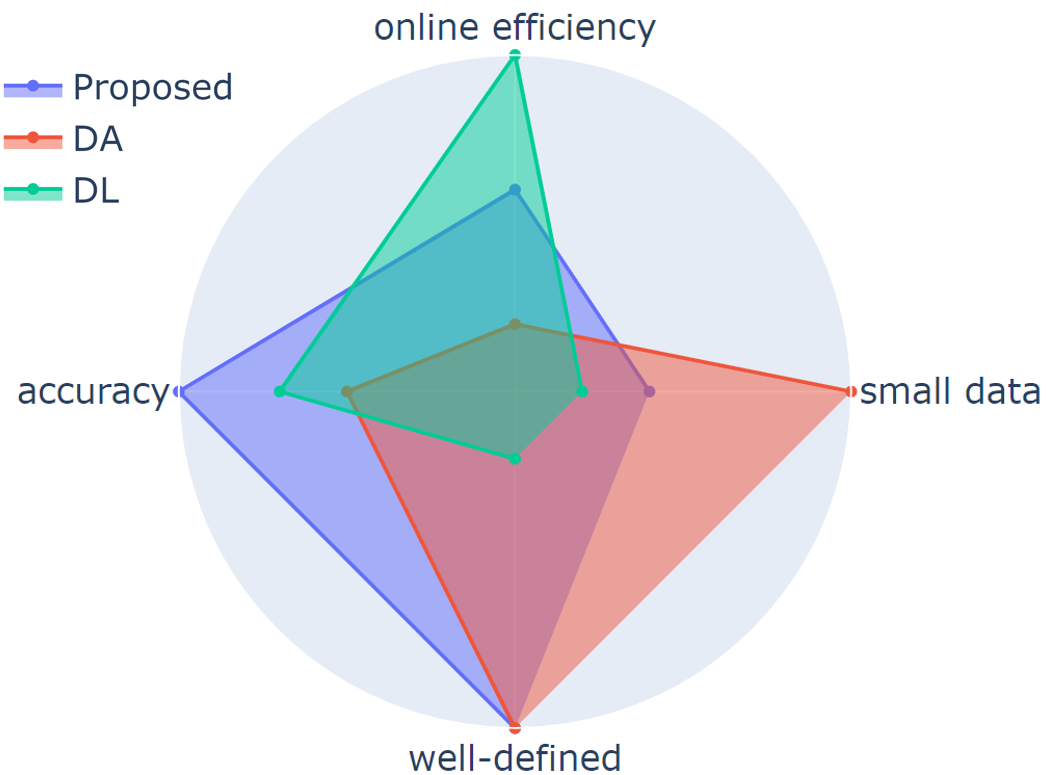}
\caption{Comparison of the proposed approach against the state-of-the-art approaches}
\label{fig:radar}
\end{figure*}

The majority of present-day deep \ac{DA} techniques, particularly those relying on \acp{CNN} or \ac{MLP}, have limitations in their ability to handle observations that are sparse, unstructured, and with varying dimensions. This poses a significant challenge when attempting to apply these methods to industrial problems. In this paper, we proposed a novel variational \ac{DA} scheme, named \ac{VIVID}, which integrates a \ac{DL} inverse operator in the assimilation objective function. By applying \ac{VCNN}, \ac{VIVID} is capable of dealing with sparse, unstructured, and time-varying sensors. In addition, the number of minimization steps in \ac{DA} can be reduced thanks to the \ac{DL} inverse operator that links directly the observations to the state space. 
We show in this paper that the proposed \ac{VIVID} can also be coupled with \ac{POD} to form an end-to-end reduced order \ac{DA} scheme to further speed-up the field reconstruction. Thorough numerical evaluations performed in the present paper demonstrate the strength of \ac{VIVID} in comparison against conventional \ac{DA} and \ac{VCNN}. As summarized in Figure~\ref{fig:radar}, \ac{VIVID} achieves high accuracy with a relatively low computational cost. It is worth noticing that by including prior estimations, \ac{VIVID} forms a well-defined problem even with sparse observations, which is not the case for most \ac{DL}-based reconstruction methods. Unlike conventional \ac{DA}, \ac{VCNN}, \ac{VIVID} and other \ac{DL} methods require offline data for training. Nevertheless, for various \ac{DA} applications, such as \ac{NWP}, hydrology or nuclear engineering, adequate historical data can be acquired from past observations.

The proposed approach can be naturally extended to include physical constraints~\cite{karniadakis2021physics} both in the \ac{DL} projection or the assimilation objective function. Future works can also consider extending \ac{VIVID} in a spatial-temporal \ac{DA} scheme (e.g., \ac{4Dvar}) where ConvLSTM~\cite{shi2015convolutional} or Transformers~\cite{vaswani2017attention} can be used to build the inverse operator for a sequence of observations. From a theoretical perspective, future work can also focus on quantifying the correlation between observation error and \ac{VCNN} error in the \ac{DA} formulation. 

\section*{Appendix: Synthetic observations in the numerical experiments}
\subsection*{Construction of the observation space}
\begin{figure*}[h!]
\centering
\includegraphics[width=0.9\textwidth]{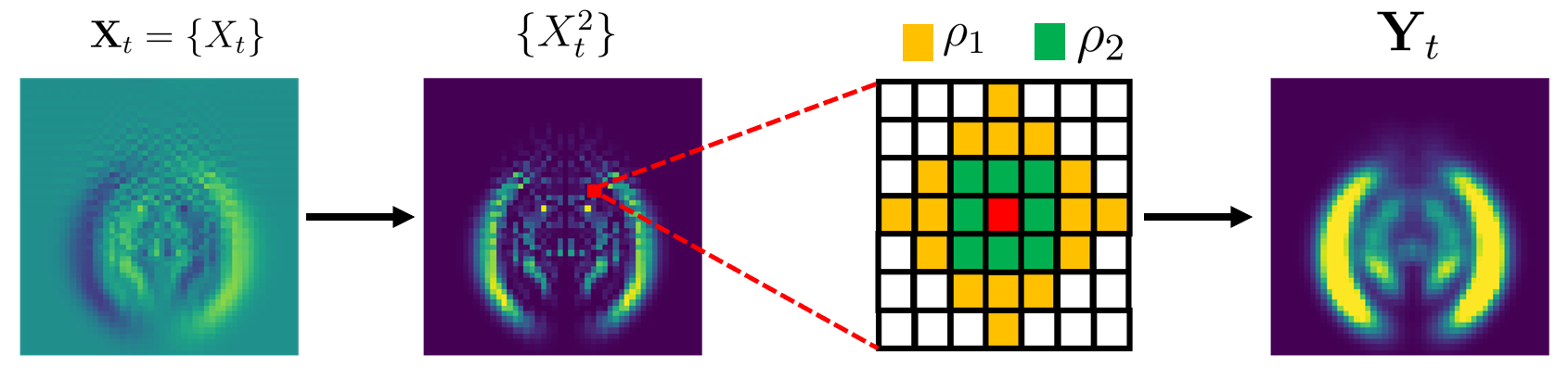}
\caption{Computation of the non-linear observation field}
\label{fig:sample.png}
\end{figure*}

In the numerical experiments of the shallow water system, the state field $\bX_t$ consists of the horizontal velocity field $u$ in Equation~\eqref{eq: sw}. The synthetic full observation field $\bY_t$ is set to be of the same dimension as $\bX_t$ (i.e., $N_x=N_y=M_x=M_y$ in this experiment). $\bY_t$ is constructed as 
\begin{align}
    \forall (i_y,j_y) & \in [1,..., M_y] \times [1,..., N_y], \notag \\
    Y_{t,i_y,j_y} &= 0.5 \sum_{(i_x,j_x) \in  \rho_1(i_y,j_y)}^{}  X^2_{t,i_x,j_x} + \sum_{(i_x,j_x) \in \rho_2(i_y,j_y)}^{} \beta_{i_x,j_x} X^2_{t,i_x,j_x}.
\end{align}
Here, 
\begin{align}
    \rho_1(i_y,j_y) &= \big\{\{i_x,j_x\} \quad | \quad \sqrt{(i_x-i_y)^2+(j_x-j_y)^2} \leq 3\big\},\\
    \rho_2(i_y,j_y) &= \big\{\{i_x,j_x\} \quad | \quad \sqrt{(i_x-i_y)^2+(j_x-j_y)^2} \leq 1.5 \big\},\\
\end{align}
representing \sibo{two neighborhoods} of different radius of the position $(i_y,j_y)$. In other words, the full synthetic observation field $\bY_t$ represents a local weighted sum of state variable squares, as shown in Figure~\ref{fig:sample.png}. 

\subsection*{Random sensor placement}
\begin{figure*}[h!]
\centering
\includegraphics[width=0.85\textwidth]{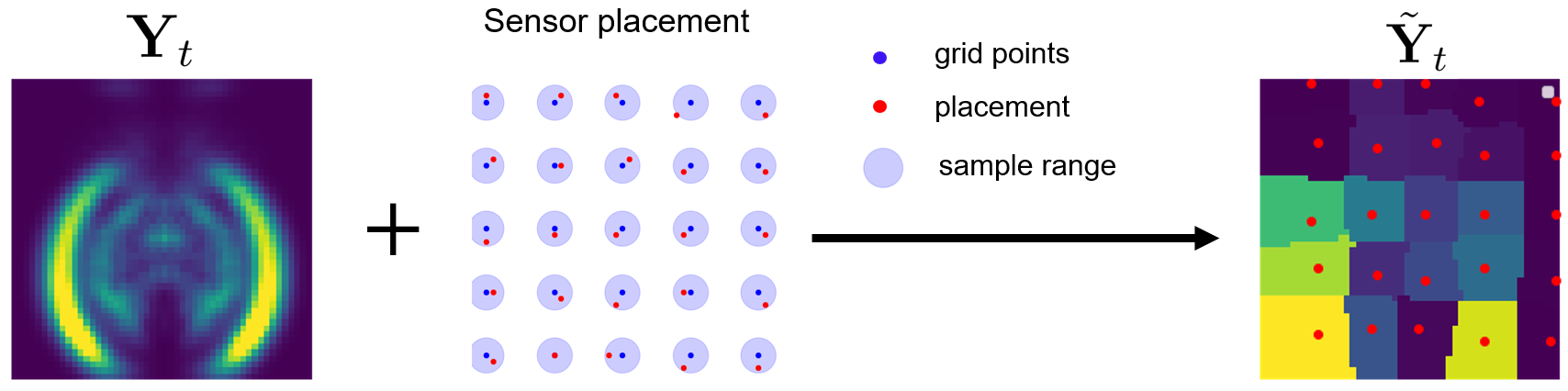}
\caption{Random sensor placement according to structured grid points}
\label{fig:sensors}
\end{figure*}

Once $\bY_t$ is computed, random sensor placement is performed based on structured grid points ($10 \times 10$ in the training set and $6 \times 6, 7 \times 7, ...,12 \times 12$ in the test set ) as shown in Figure~\ref{fig:sensors}. More precisely, a sensor is randomly positioned in the neighbourhood of each grid point within a sample range $r_s = 3$. Thus the observations $\{y_{t,k}\}_{k=1...k^*} $ and the observation vector $\by_t$ are differentiable with respect to the state $\bX_t$.\\

\section*{Data and code availability}
The code of the shallow water experiments is available at \url{https://github.com/DL-WG/VIVID}. Sample data and the script to generate experimental is also provided in the github reporsitory. 

\section*{acknowledgement}

\sibo{We thank the editor and the anonymous reviewer for their careful reading of our manuscript and their many insightful comments and suggestions.} This work is supported by the Leverhulme Centre for Wildfires, Environment and Society through the Leverhulme Trust, grant number RC-2018-023 and the EP/T000414/1 PREdictive Modelling with
Quantification of UncERtainty for MultiphasE Systems (PREMIERE).

\section*{Acronyms}

\begin{acronym}[AAAAA]
\footnotesize{
\acro{NN}{Neural Network}
\acro{ML}{Machine Learning}
\acro{LA}{Latent Assimilation}
\acro{DA}{Data Assimilation}
\acro{AE}{Autoencoder}
\acro{CAE}{Convolutional Autoencoder}
\acro{BLUE}{Best Linear Unbiased Estimator}
\acro{RNN}{Recurrent Neural Network}
\acro{CNN}{Convolutional Neural Network}
\acro{SSIM}{Structural Similarity Index Measure}
\acro{LSTM}{long short-term memory}
\acro{POD}{Proper Orthogonal Decomposition}
\acro{VIVID}{Voronoi-tessellation Inverse operator for VariatIonal Data assimilation}
\acro{VCNN}{Voronoi-tessellation Convolutional Neural Network}
\acro{SVD}{Singular Value Decomposition}
\acro{ROM}{reduced-order modelling}
\acro{CFD}{Computational Fluid Dynamics}
\acro{NWP}{Numerical Weather Prediction}
\acro{R-RMSE}{Relative Root Mean Square Error}
\acro{BFGS}{Broyden–Fletcher–Goldfarb–Shanno}
\acro{AI}{artificial intelligence}
\acro{DL}{Deep Learning}
\acro{KF}{Kalman filter}
\acro{MLP}{Multi Layer Percepton}
\acro{GNN}{Graph Neural Network}
\acro{GLA}{Generalised Latent Assimilation}
\acro{3Dvar}{Three-dimensional  variational data assimilation }
\acro{4Dvar}{Four-dimensional  variational data assimilation }}
\end{acronym}

\footnotesize{
\bibliographystyle{elsarticle-num-names}
\bibliography{main.bib}}
\end{document}